\documentclass[nohyperref]{article}

\usepackage[
	activate={true,nocompatibility},
	tracking=true
]{microtype}
\usepackage{graphicx}
\usepackage{subfigure}
\usepackage{booktabs} % for professional tables
\usepackage{amsmath}
\usepackage{amssymb}
\usepackage{mathtools}
\usepackage{makecell}
\usepackage{multirow}
\usepackage{comment}

\newcommand{\modelname}{Point$\cdot$E}

\makeatletter
\def\blfootnote{\xdef\@thefnmark{}\@footnotetext}
\makeatother

\usepackage[accepted]{icml2022}

\usepackage{hyperref}
\usepackage[all]{hypcap}

\icmltitlerunning{\modelname{}: A System for Generating 3D Point Clouds from Complex Prompts}

\begin{document}

\twocolumn[
\icmltitle{\modelname{}: A System for Generating 3D Point Clouds from Complex Prompts}



\icmlsetsymbol{equal}{*}

\begin{icmlauthorlist}
\icmlauthor{Alex Nichol}{equal,op}
\icmlauthor{Heewoo Jun}{equal,op}
\icmlauthor{Prafulla Dhariwal}{op}
\icmlauthor{Pamela Mishkin}{op}
\icmlauthor{Mark Chen}{op}
\end{icmlauthorlist}

\icmlaffiliation{op}{OpenAI, San Francisco, USA}

\icmlcorrespondingauthor{Alex Nichol}{alex@openai.com}
\icmlcorrespondingauthor{Heewoo Jun}{heewoo@openai.com}

\icmlkeywords{Machine Learning, image generation, diffusion models, guided diffusion, CLIP}

\vskip 0.3in
]



\printAffiliationsAndNotice{\icmlEqualContribution} % otherwise use the standard text.

\begin{abstract}
While recent work on text-conditional 3D object generation has shown promising results, the state-of-the-art methods typically require multiple GPU-hours to produce a single sample. This is in stark contrast to state-of-the-art generative image models, which produce samples in a number of seconds or minutes. In this paper, we explore an alternative method for 3D object generation which produces 3D models in only 1-2 minutes on a single GPU. Our method first generates a single synthetic view using a text-to-image diffusion model, and then produces a 3D point cloud using a second diffusion model which conditions on the generated image. While our method still falls short of the state-of-the-art in terms of sample quality, it is one to two orders of magnitude faster to sample from, offering a practical trade-off for some use cases. We release our pre-trained point cloud diffusion models, as well as evaluation code and models, at \url{https://github.com/openai/point-e}.
\end{abstract}

\section{Introduction}
\label{sec:introduction}

\begin{figure*}[ht!]
    \centering
    \includegraphics[width=0.94\textwidth]{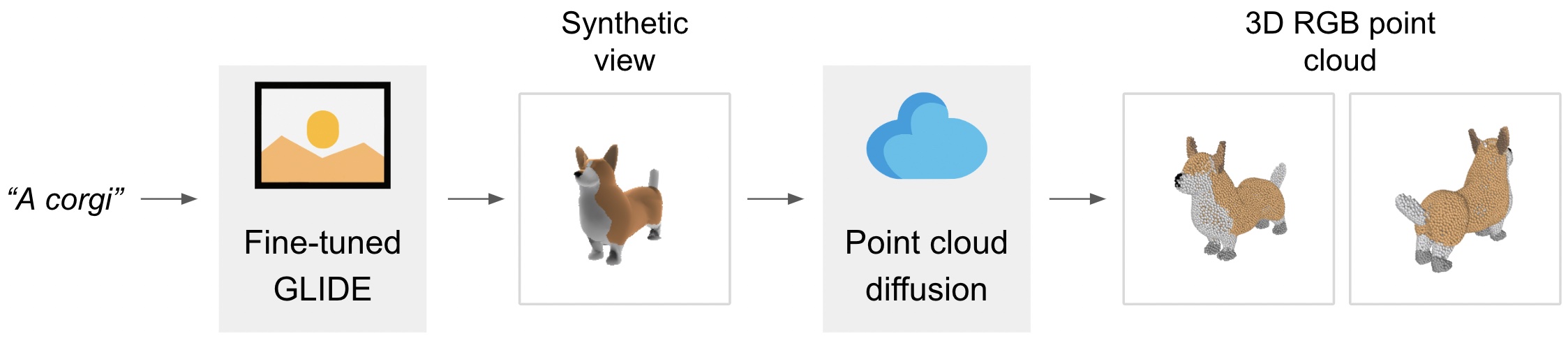}
    \caption{\label{fig:hero} A high-level overview of our pipeline. First, a text prompt is fed into a GLIDE model to produce a synthetic rendered view. Next, a point cloud diffusion stack conditions on this image to produce a 3D RGB point cloud.}
    \vskip -0.1in
\end{figure*}

\begin{figure*}[h!]
    \centering
    \setlength{\tabcolsep}{2.0pt}
    \begin{tabular}{cccc}
        \includegraphics[width=0.25\textwidth]{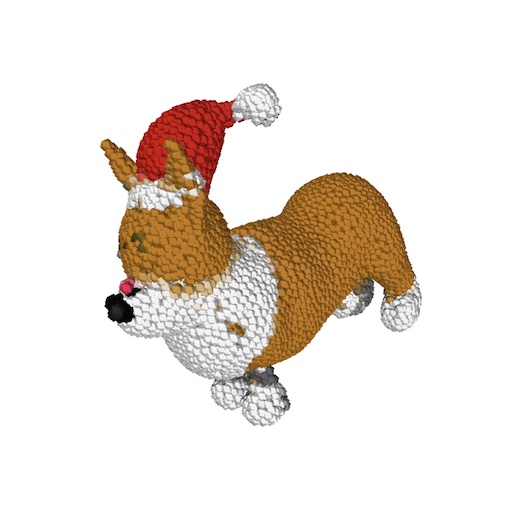} &
        \includegraphics[width=0.25\textwidth]{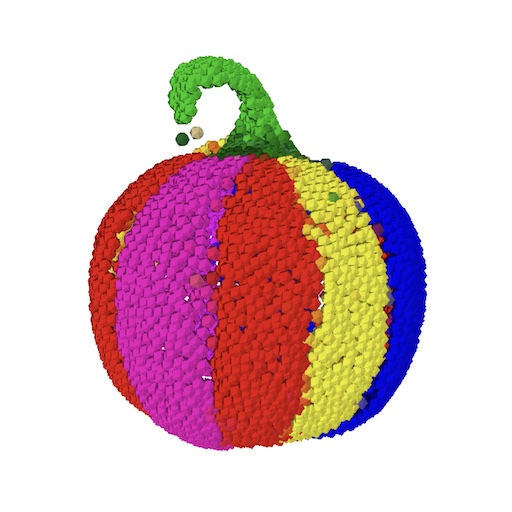} &
        \includegraphics[width=0.25\textwidth]{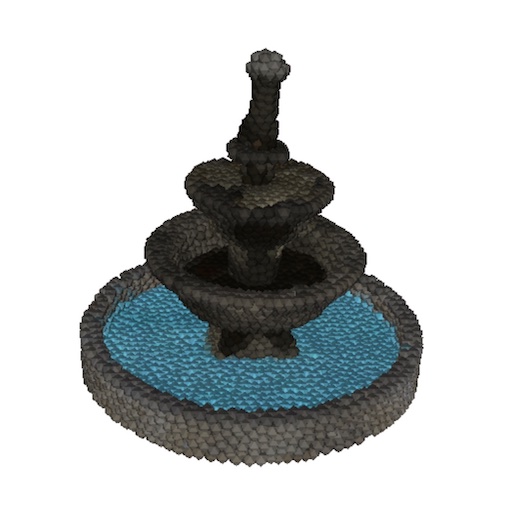} &
        \includegraphics[width=0.25\textwidth]{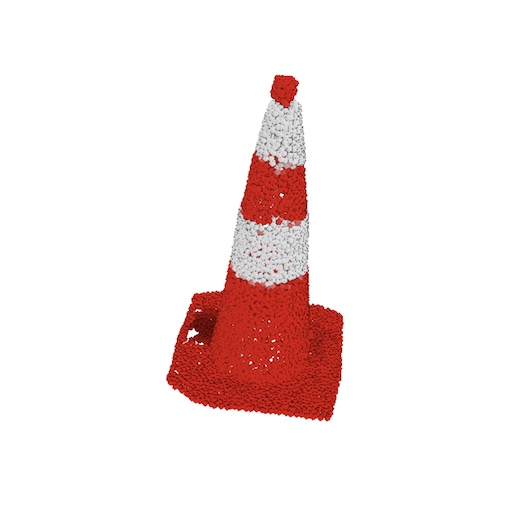} \\

        \scriptsize \makecell{``a corgi wearing a \\ red santa hat''} &
        \scriptsize \makecell{``a multicolored rainbow \\ pumpkin''} &
        \scriptsize \makecell{``an elaborate fountain''} &
        \scriptsize \makecell{``a traffic cone''} \\
        \rule{0pt}{0.15pt} \\
        
        \includegraphics[width=0.25\textwidth]{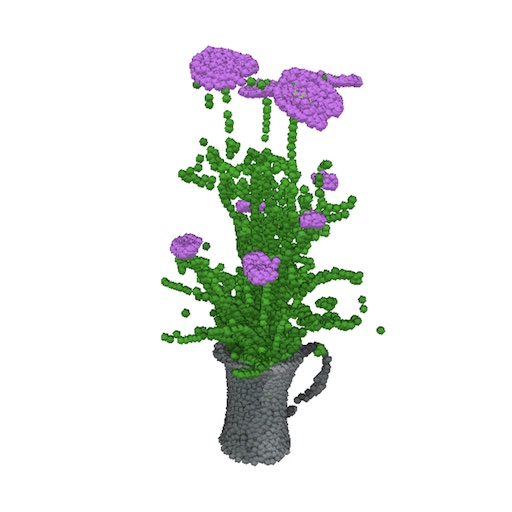} &
        \includegraphics[width=0.25\textwidth]{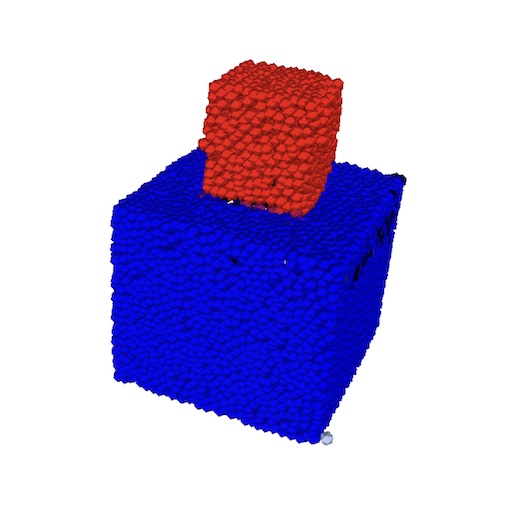} &
        \includegraphics[width=0.25\textwidth]{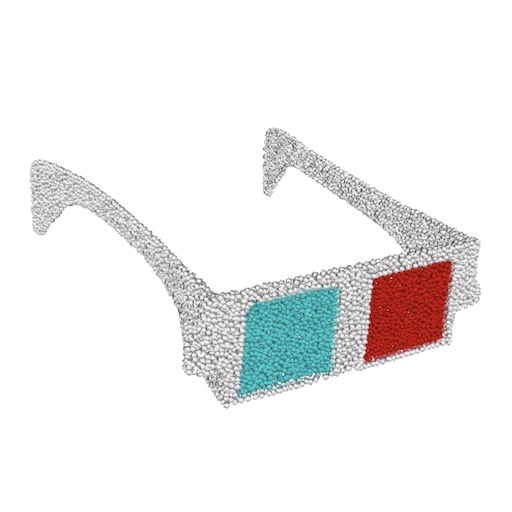} &
        \includegraphics[width=0.25\textwidth]{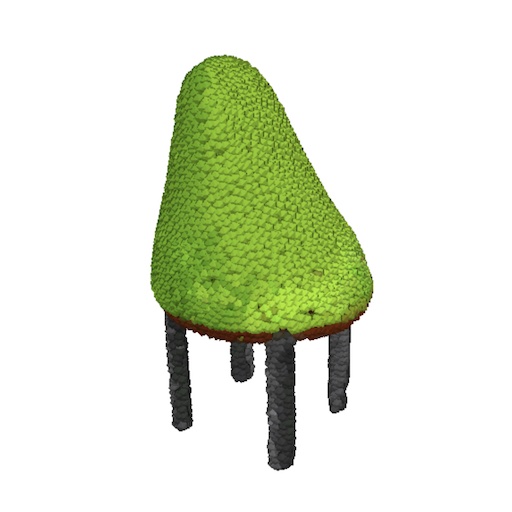} \\

        \scriptsize \makecell{``a vase of purple flowers''} &
        \scriptsize \makecell{``a small red cube is sitting \\ on top of a large blue cube. \\ red on top, blue on bottom''} &
        \scriptsize \makecell{``a pair of 3d glasses, \\ left lens is red right \\ is blue''} &
        \scriptsize \makecell{``an avocado chair, a chair \\ imitating an avocado''} \\
        \rule{0pt}{0.15pt} \\

        \includegraphics[width=0.25\textwidth]{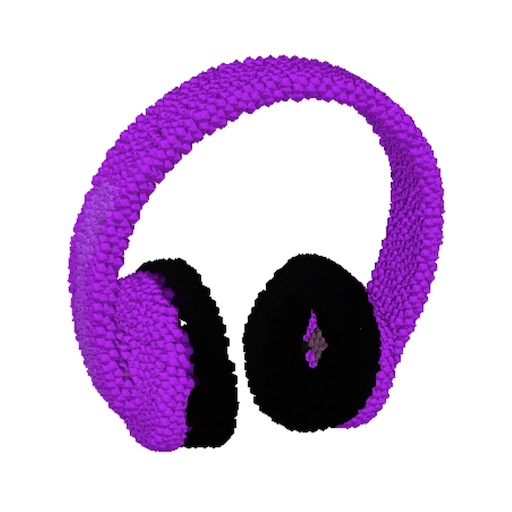} &
        \includegraphics[width=0.25\textwidth]{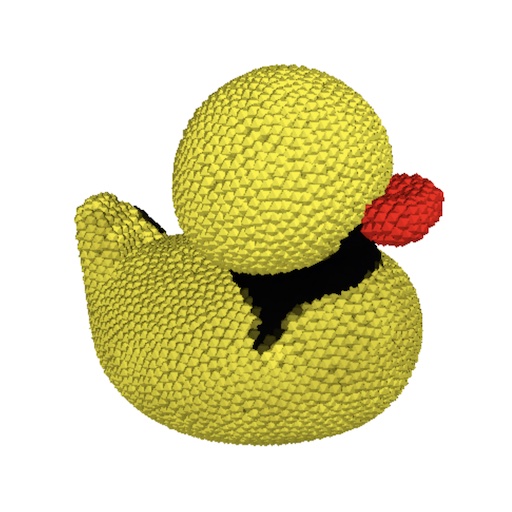} &
        \includegraphics[width=0.25\textwidth]{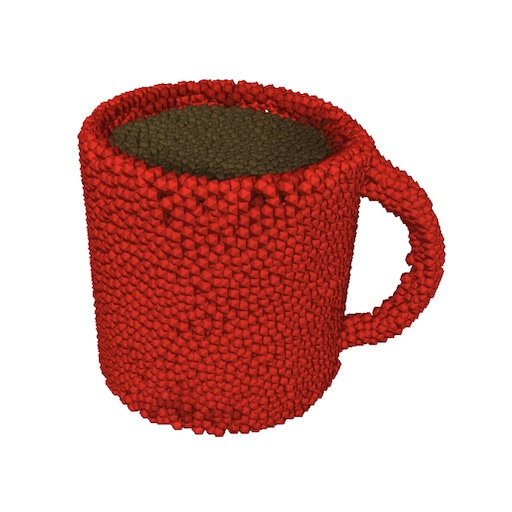} &
        \includegraphics[width=0.25\textwidth]{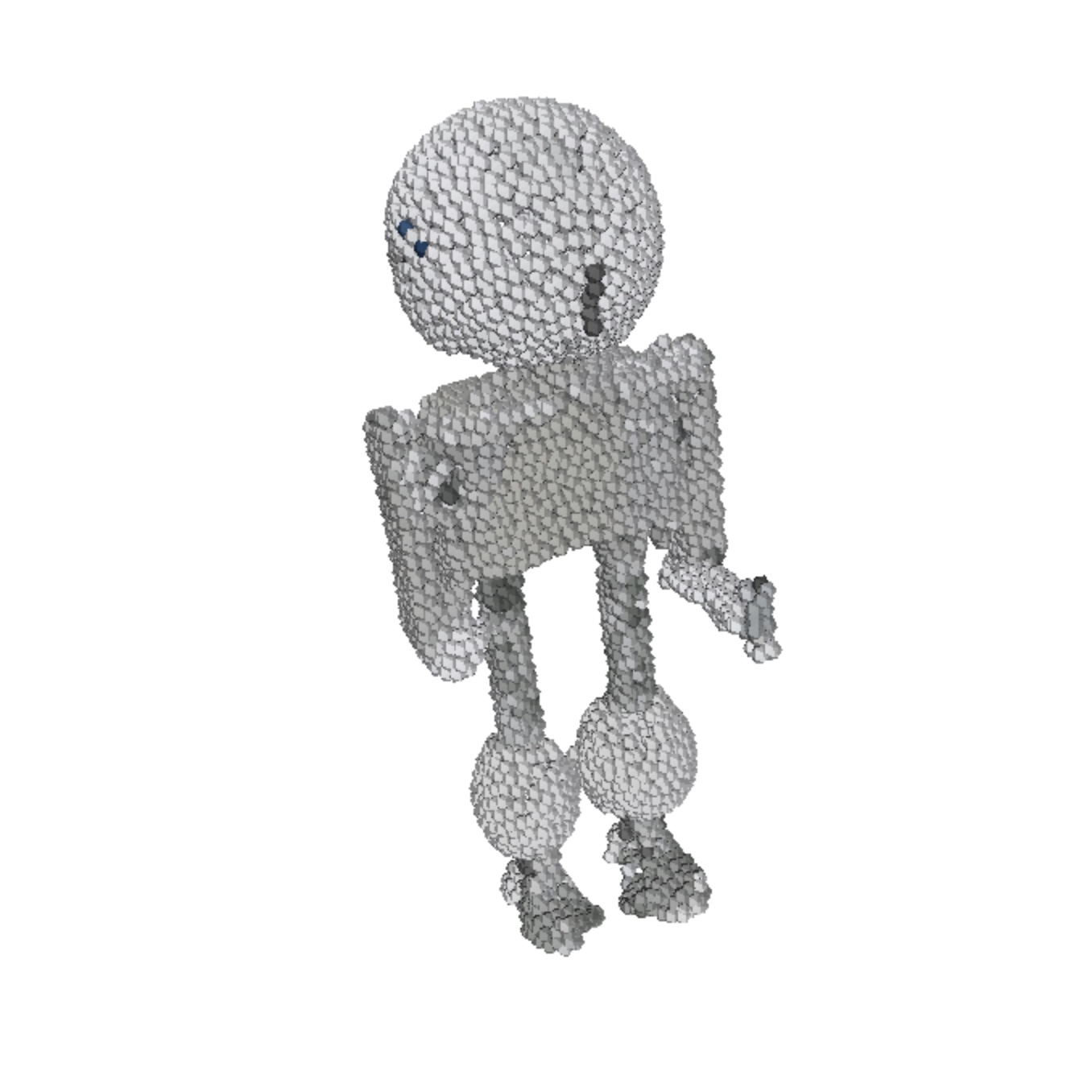} \\

        \scriptsize \makecell{``a pair of purple \\ headphones''} &
        \scriptsize \makecell{``a yellow rubber duck''} &
        \scriptsize \makecell{``a red mug filled \\ with coffee''} &
        \scriptsize \makecell{``a humanoid robot with \\ a round head''} \\
        \rule{0pt}{0.15pt} \\
    \end{tabular}

    \caption{Selected point clouds generated by \modelname{} using the given text prompts. For each prompt, we selected one point cloud out of eight samples.}
    \label{fig:banner_images}
    \vskip -0.1in
\end{figure*}

With the recent explosion of text-to-image generative models, it is now possible to generate and modify high-quality images from natural language descriptions in a number of seconds \citep{dalle,cogview,glide,unclip,makeascene,parti,imagen,engievilg2,ediffi}. Inspired by these results, recent works have explored text-conditional generation in other modalities, such as video \citep{cogvideo,makeavideo,videodm,imagenvideo} and 3D objects \citep{dreamfields,dreamfusion,magic3d,clipforge,textcraft}. In this work, we focus specifically on the problem of text-to-3D generation, which has significant potential to democratize 3D content creation for a wide range of applications such as virtual reality, gaming, and industrial design.

Recent methods for text-to-3D synthesis typically fall into one of two categories:

\begin{enumerate}
  \item Methods which train generative models directly on paired \mbox{(text, 3D)} data \citep{text2shape,autosdf,shapecrafter,lion} or unlabeled 3D data \citep{clipforge,textcraft,3dim}. While these methods can leverage existing generative modeling approaches to produce samples efficiently, they are difficult to scale to diverse and complex text prompts due to the lack of large-scale 3D datasets \citep{textcraft}.
  \item Methods which leverage pre-trained text-image models to optimize differentiable 3D representations \citep{dreamfields,dreamfusion,magic3d}. These methods are often able to handle complex and diverse text prompts, but require expensive optimization processes to produce each sample. Furthermore, due to the lack of a strong 3D prior, these methods can fall into local minima which don't correspond to meaningful or coherent 3D objects \citep{dreamfusion}.
\end{enumerate}

We aim to combine the benefits of both categories by pairing a text-to-image model with an image-to-3D model. Our text-to-image model leverages a large corpus of \mbox{(text, image)} pairs, allowing it to follow diverse and complex prompts, while our image-to-3D model is trained on a smaller dataset of \mbox{(image, 3D)} pairs. To produce a 3D object from a text prompt, we first sample an image using the text-to-image model, and then sample a 3D object conditioned on the sampled image. Both of these steps can be performed in a number of seconds, and do not require expensive optimization procedures. Figure \ref{fig:hero} depicts this two-stage generation process.

We base our generative stack on diffusion \citep{dickstein,scorematching,ddpm}, a recently proposed generative framework which has become a popular choice for text-conditional image generation.
For our text-to-image model, we use a version of GLIDE \citep{glide} fine-tuned on 3D renderings (Section \ref{sec:singleviewglide}). For our image-to-3D model, we use a stack of diffusion models which generate RGB point clouds conditioned on images (Section \ref{sec:basediffusion} and \ref{sec:upsampler} detail our novel Transformer-based architecture for this task). For rendering-based evaluations, we go one step further and produce meshes from generated point clouds using a regression-based approach (Section \ref{sec:mesh}).

We find that our system can often produce colored 3D point clouds that match both simple and complex text prompts (See Figure \ref{fig:banner_images}). We refer to our system as \modelname{}, since it generates \underline{\textbf{point}} clouds \underline{\textbf{e}}fficiently. We release our point cloud diffusion models, as well as evaluation code and models, at \url{https://github.com/openai/point-e}.

\section{Background}
\label{sec:background}

Our method builds off of a growing body of work on diffusion-based models, which were first proposed by \citet{dickstein} and popularized more recently \citep{scorematching,improvedscore,ddpm}.

We follow the Gaussian diffusion setup of \citet{ddpm}, which we briefly describe here. We aim to sample from some distribution $q(x_0)$ using a neural network approximation $p_{\theta}(x_0)$. Under Gaussian diffusion, we define a noising process 

$$q(x_t|x_{t-1}) \coloneqq \mathcal{N}(x_t; \sqrt{1-\beta_t} x_{t-1}, \beta_t \mathbf{I})$$

for integer timesteps $t \in [0, T]$. Intuitively, this process gradually adds Gaussian noise to a signal, with the amount of noise added at each timestep determined by some noise schedule $\beta_t$. We employ a noise schedule such that, by the final timestep $t=T$, the sample $x_T$ contains almost no information (i.e. it looks like Gaussian noise). \citet{ddpm} note that it is possible to directly jump to a given timestep of the noising process without running the whole chain:

$$x_t = \sqrt{\bar{\alpha}_t} x_0 + \sqrt{1-\bar{\alpha}_t} \epsilon$$

where $\epsilon \sim \mathcal{N}(0, \mathbf{I})$ and $\bar{\alpha}_t \coloneqq \prod_{s=0}^t 1-\beta_t$.

To train a diffusion model, we approximate $q(x_{t-1}|x_t)$ as a neural network $p_{\theta}(x_{t-1}|x_t)$. We can then produce a sample by starting at random Gaussian noise $x_T$ and gradually reversing the noising process until arriving at a noiseless sample $x_0$. With enough small steps, $p_{\theta}(x_{t-1}|x_t)$ can be parameterized as a diagonal Gaussian distribution, and \citet{ddpm} propose to parameterize the mean of this distribution by predicting $\epsilon$, the effective noise added to a sample $x_t$. While \citet{ddpm} fix the variance $\Sigma$ of $p_{\theta}(x_{t-1}|x_t)$ to a reasonable per-timestep heuristic, \citet{improved} achieve better results by predicting the variance as well as the mean.

Diffusion sampling can be cast through the lens of differential equations \citep{sde}, allowing one to use various SDE and ODE solvers to sample from these models. \citet{edm} find that a carefully-designed second-order ODE solver provides a good trade-off between quality and sampling efficiency, and we employ this sampler for our point cloud diffusion models.

To trade off sample diversity for fidelity in diffusion models, several \textit{guidance} strategies may be used. \citet{sotapaper} introduce classifier guidance, where gradients from a noise-aware classifier $\nabla_{x_t} p_{\theta}(y|x_t)$ are used to perturb every sampling step. They find that increasing the scale of the perturbation increases generation fidelity while reducing sample diversity. \citet{uncond} introduce classifier-free guidance, wherein a conditional diffusion model $p(x_{t-1}|x_t, y)$ is trained with the class label stochastically dropped and replaced with an additional $\varnothing$ class. During sampling, the model's output $\epsilon$ is linearly extrapolated away from the unconditional prediction towards the conditional prediction:

$$\epsilon_{\textrm{guided}} \coloneqq \epsilon_\theta(x_t, \varnothing) + s \cdot (\epsilon_\theta(x_t, y) - \epsilon_\theta(x_t, \varnothing))$$

for some guidance scale $s \ge 1$. This approach is straightforward to implement, requiring only that conditioning information is randomly dropped during training time. We employ this technique throughout our models, using the drop probability 0.1.

\section{Related Work}
\label{sec:related_work}

Several prior works have explored generative models over point clouds. \citet{pcautoencoder} train point cloud auto-encoders, and fit generative priors (either GANs \citep{gan} or GMMs) on the resulting latent representations. \citet{structurenet} generate point clouds using a VAE \citep{vae} on hierarchical graph representations of 3D objects. \citet{pointflow} train a two-stage flow model for point cloud generation: first, a prior flow model produces a latent vector, and then a second flow model samples points conditioned on the latent vector. Along the same lines, \citet{flowdiff,gradientfields} both train two-stage models where the second stage is a diffusion model over individual points in a point cloud, and the first stage is a latent flow model or a latent GAN, respectively. \citet{lion} train a two-stage hierarchical VAE on point clouds with diffusion priors at both stages. Most similar to our work, \citet{pvd} introduce PVD, a single diffusion model that generates point clouds directly. Compared to previous point cloud diffusion methods such as PVD, our Transformer-based model architecture is simpler and incorporates less 3D-specific structure. Unlike prior works, our models also produce RGB channels alongside point cloud coordinates.

A growing body of work explores the problem of 3D model generation in representations other than point clouds. Several works aim to train 3D-aware GANs from datasets of 2D images \citep{pigan, graf, eg3d, stylesdf, stylenerf, cips3d}. These GANs are typically applied to the problem of novel view synthesis in forward-facing scenes, and do not attempt to reconstruct full 360-degree views of objects. More recently, \citet{get3d} train a GAN that directly produces full 3D meshes, paired with a discriminator that inputs differentiably-rendered \citep{nvdiffrast} views of the generated meshes. \citet{gaudi} generates complete 3D scenes by first learning a representation space that decodes into NeRFs \citep{nerf}, and then training a diffusion prior on this representation space. However, none of these works have demonstrated the ability to generate arbitrary 3D models conditioned on open-ended, complex text-prompts.

Several recent works have explored the problem of text-conditional 3D generation by optimizing 3D representations according to a text-image matching objective. \citet{dreamfields} introduce DreamFields, a method which optimizes the parameters of a NeRF using an objective based on CLIP \citep{clip}. Notably, this method requires no 3D training data. Building on this principle, \citet{clipmesh} optimizes a mesh using a CLIP-guided objective, finding that the mesh representation is more efficient to optimize than a NeRF. More recently, \citet{dreamfusion} extend DreamFields to leverage a pre-trained text-to-image diffusion model instead of CLIP, producing more coherent and complex objects. \citet{magic3d} build off of this technique, but convert the NeRF representation into a mesh and then refine the mesh representation in a second optimization stage. While these approaches are able to produce diverse and complex objects or scenes, the optimization procedures typically require multiple GPU hours to converge, making them difficult to apply in practical settings.

While the above approaches are all based on optimization against a text-image model and do not leverage 3D data, other methods for text-conditional 3D synthesis make use of 3D data, possibly paired with text labels. \citet{text2shape} employ a dataset of text-3D pairs to train a GAN to generate 3D representations conditioned on text. \citet{towardsimplicit} also leverage paired text-3D data to generate models in a joint representation space. \citet{clipforge} employ a flow-based model to generate 3D latent representations, and find some text-to-3D capabilities when conditioning their model on CLIP embeddings. More recently, \citet{lion} achieve similar results when conditioning on CLIP embeddings, but employ a hierarchical VAE on point clouds for their generative stack. \citet{autosdf} and \citet{shapecrafter} employ a VQ-VAE \citep{vqvae} with an autoregressive prior to sample 3D shapes conditioned on text labels. More recently, \citet{textcraft} also employ a VQ-VAE approach, but leverage CLIP embeddings to avoid the need for explicit text labels in the dataset. While many of these works demonstrate promising early results, they tend to be limited to simple prompts or a narrow set of object categories due to the limited availability of 3D training data. Our method sidesteps this issue by leveraging a pre-trained text-to-image model to condition our 3D generation procedure.

A large body of research focuses on reconstructing 3D models from single or few images. Notably, this is an underspecified problem, since the model must impute some details not present in the conditioning image(s). Nevertheless, some regression-based methods have shown promising results on this task \citep{r2n2,pixel2mesh,meshrcnn,atlasnet,pixelnerf,visionnerf}. A separate body of literature studies generative approaches for single- or multi-view reconstruction. \citet{pointset} predict point clouds of objects from single views using a VAE. \citet{multiviewnovel} use a hybrid of a flow predictor and a GAN to generate novel views from few images. \citet{nerfvae} use a view-conditional VAE to generate latent vectors for a NeRF decoder. \citet{3dim} employ an image-to-image diffusion model to synthesize novel views of an object conditioned on a single view, allowing many consistent views to be synthesized autoregressively.

\section{Method}

%We aim to model $p(x|\textrm{text})$, the distribution of 3D point clouds $x$ given a text caption.
Rather than training a single generative model to directly produce point clouds conditioned on text, we instead break the generation process into three steps. First, we generate a synthetic view conditioned on a text caption. Next, we produce a coarse point cloud (1,024 points) conditioned on the synthetic view. And finally, we produce a fine point cloud (4,096 points) conditioned on the low-resolution point cloud and the synthetic view. In practice, we assume that the image contains the relevant information from the text, and do not explicitly condition the point clouds on the text.
%In particular, we factorize the distribution using the chain rule:
%
% \begin{equation*}
%   \begin{aligned}
%     p(x|\textrm{text}) = & \textrm{ } p_{\textrm{glide}}(\textrm{image}|\textrm{text}) \\
%     & \cdot p_{\textrm{base}}(\textrm{coarse PC} | \textrm{image},\textrm{text}) \\
%     & \cdot p_{\textrm{upsampler}}(\textrm{fine PC} | \textrm{coarse PC}, \textrm{image},\textrm{text})
%   \end{aligned}
% \end{equation*}

% $$p(x|\textrm{text}) = p_v(x_v|\textrm{text}) \cdot p_c(x_c | x_v,\textrm{text}) \cdot p_f(x | x_c, x_v,\textrm{text})$$

%where $p_v(x_v|\textrm{text})$ is a text-to-image model which samples an image $x_v$, $p_c(x_c|x_v,\textrm{text})$ is a point cloud diffusion model which samples a coarse point cloud $x_c$, and finally $p_f(x|x_c,x_v,\textrm{text})$ is a second point cloud diffusion model which samples a final point cloud $x$.

%In practice, we assume that the image contains the relevant information from the text, and do not explicitly condition $p_{\textrm{base}}h$ or $p_{\textrm{upsampler}}$ on the text.

To generate text-conditional synthetic views, we use a 3-billion parameter GLIDE model \citep{glide} fine-tuned on rendered 3D models from our dataset (\mbox{Section \ref{sec:singleviewglide}}). To generate low-resolution point clouds, we use a conditional, permutation invariant diffusion model (Section \ref{sec:basediffusion}). To upsample these low-resolution point clouds, we use a similar (but smaller) diffusion model which is additionally conditioned on the low-resolution point cloud (Section \ref{sec:upsampler}).

We train our models on a dataset of several million 3D models and associated metadata. We process the dataset into rendered views, text descriptions, and 3D point clouds with associated RGB colors for each point. We describe our data processing pipeline in more detail in Section \ref{sec:dataset}.

\subsection{Dataset}
\label{sec:dataset}

We train our models on several million 3D models. We found that data formats and quality varied wildly across our dataset, prompting us to develop various post-processing steps to ensure higher data quality.

To convert all of our data into one generic format, we rendered every 3D model from 20 random camera angles as RGBAD images using Blender \citep{blender}, which supports a variety of 3D formats and comes with an optimized rendering engine. For each model, our Blender script normalizes the model to a bounding cube, configures a standard lighting setup, and finally exports RGBAD images using Blender's built-in realtime rendering engine.

We then converted each object into a colored point cloud using its renderings. In particular, we first constructed a dense point cloud for each object by computing points for each pixel in each RGBAD image. These point clouds typically contain hundreds of thousands of unevenly spaced points, so we additionally used farthest point sampling to create uniform clouds of 4K points. By constructing point clouds directly from renders, we were able to sidestep various issues that might arise from attempting to sample points directly from 3D meshes, such as sampling points which are contained within the model or dealing with 3D models that are stored in unusual file formats.

Finally, we employed various heuristics to reduce the frequency of low-quality models in our dataset. First, we eliminated flat objects by computing the SVD of each point cloud and only retaining those where the smallest singular value was above a certain threshold. Next, we clustered the dataset by CLIP features (for each object, we averaged features over all renders). We found that some clusters contained many low-quality categories of models, while other clusters appeared more diverse or interpretable. We binned these clusters into several buckets of varying quality, and used a weighted mixture of the resulting buckets as our final dataset.

\subsection{View Synthesis GLIDE Model}
\label{sec:singleviewglide}

Our point cloud models are conditioned on rendered views from our dataset, which were all produced using the same renderer and lighting settings. Therefore, to ensure that these models correctly handle generated synthetic views, we aim to explicitly generate 3D renders that match the distribution of our dataset.

To this end, we fine-tune GLIDE with a mixture of its original dataset and our dataset of 3D renderings. Since our 3D dataset is small compared to the original GLIDE training set, we only sample images from the 3D dataset 5\% of the time, using the original dataset for the remaining 95\%. We fine-tune for 100K iterations, meaning that the model has made several epochs over the 3D dataset (but has never seen the same exact rendered viewpoint twice).

To ensure that we always sample in-distribution renders (rather than only sampling them 5\% of the time), we add a special token to every 3D render's text prompt indicating that it is a 3D render; we then sample with this token at test time.

\subsection{Point Cloud Diffusion}
\label{sec:basediffusion}

\begin{figure}[t]
    \centering
    \includegraphics[width=0.45\textwidth]{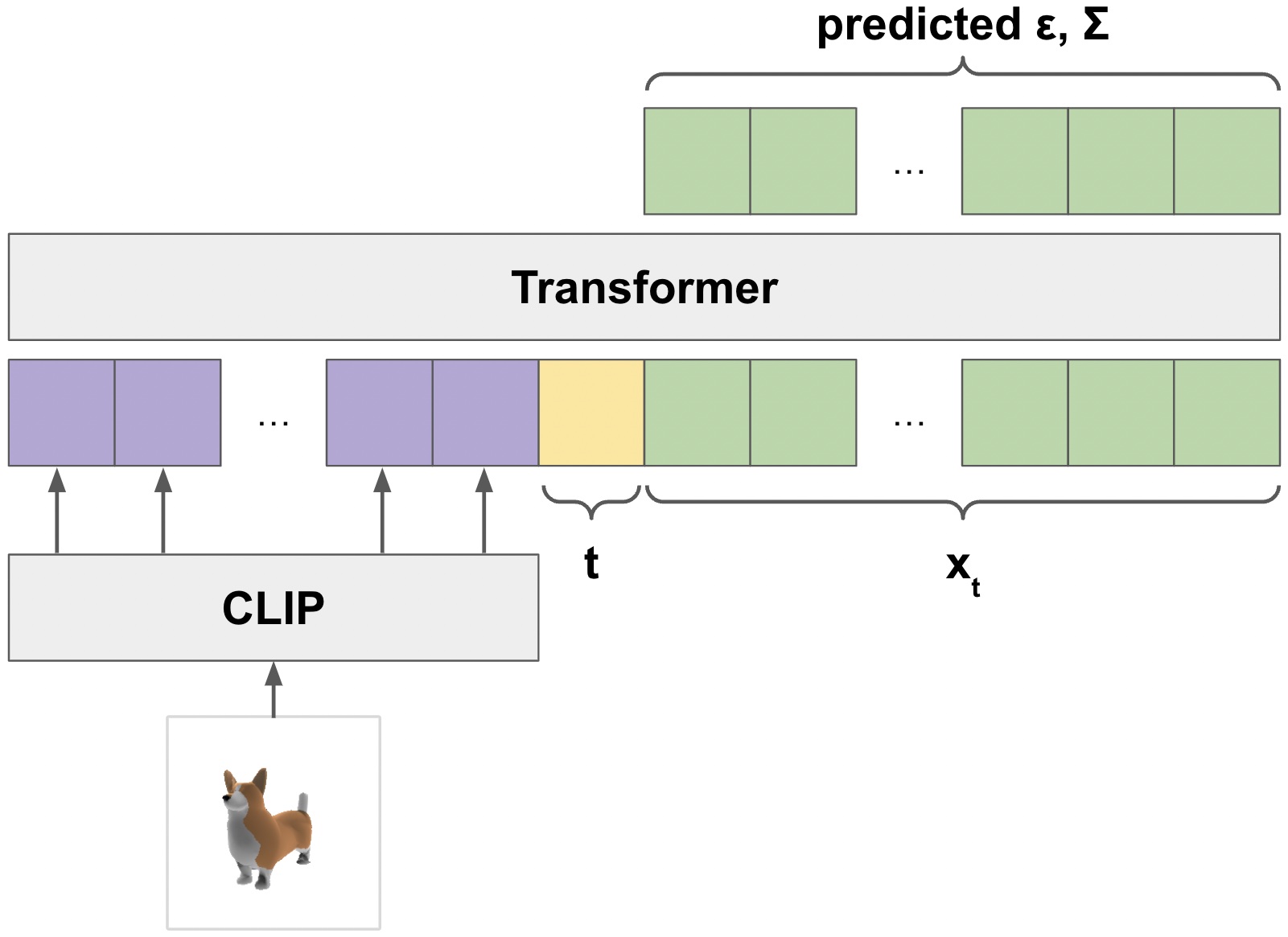}
    \caption{\label{fig:architecture} Our point cloud diffusion model architecture. Images are fed through a frozen, pre-trained CLIP model, and the output grid is fed as tokens into the transformer. Both the timestep $t$ and noised input $x_t$ are also fed in as tokens. The output tokens corresponding to $x_t$ are used to predict $\epsilon$ and $\Sigma$.}
    \vskip -0.1in
\end{figure}

To generate point clouds with diffusion, we extend the framework used by \citet{pvd} to include RGB colors for each point in a point cloud. In particular, we represent a point cloud as a tensor of shape $K \times 6$, where $K$ is the number of points, and the inner dimension contains $(x,y,z)$ coordinates as well as $(R, G, B)$ colors. All coordinates and colors are normalized to the range $[-1, 1]$. We then generate these tensors directly with diffusion, starting from random noise of shape $K \times 6$, and gradually denoising it.

Unlike prior work which leverages 3D-specific architectures to process point clouds, we use a simple Transformer-based model \citep{transformer} to predict both $\epsilon$ and $\Sigma$ conditioned on the image, timestep $t$, and noised point cloud $x_t$. An overview of our architecture can be seen in Figure \ref{fig:architecture}. As input context to this model, we run each point in the point cloud through a linear layer with output dimension $D$, obtaining a $K \times D$ input tensor. Additionally, we run the timestep $t$ through a small MLP, obtaining another $D$-dimensional vector to prepend to the context. 

To condition on the image, we feed it through a pre-trained ViT-L/14 CLIP model, take the last layer embeddings from this CLIP model (of shape $256 \times D'$), and linearly project it into another tensor of shape $256 \times D$ before prepending it to the Transformer context. In Section \ref{sec:modelablations}, we find that this is superior to using a single CLIP image or text embedding, as done by \citet{clipforge,lion,textcraft}.

The final input context to our model is of shape $(K+257) \times D$. To obtain a final output sequence of length $K$, we take the final $K$ tokens of output and project it to obtain $\epsilon$ and $\Sigma$ predictions for the $K$ input points.

Notably, we do not employ positional encodings for this model. As a result, the model itself is permutation-invariant to the input point clouds (although the output order is tied to the input order).

\subsection{Point Cloud Upsampler}
\label{sec:upsampler}

For image diffusion models, the best quality is typically achieved by using some form of hierarchy, where a low-resolution base model produces output which is then upsampled by another model \citep{improved,sr3,cascaded,latentdiffusion}. We employ this approach to point cloud generation by first generating 1K points with a large base model, and then upsampling to 4K points using a smaller upsampling model. Notably, our models' compute requirements scale with the number of points, so it is four times more expensive to generate 4K points than 1K points for a fixed model size.

Our upsampler uses the same architecture as our base model, with extra conditioning tokens for the low-resolution point cloud. To arrive at 4K points, the upsampler conditions on 1K points and generates an additional 3K points which are added to the low-resolution pointcloud. We pass the conditioning points through a separate linear embedding layer than the one used for $x_t$, allowing the model to distinguish conditioning information from new points without requiring the use of positional embeddings.

\subsection{Producing Meshes}
\label{sec:mesh}

For rendering-based evaluations, we do not render generated point clouds directly. Rather, we convert the point clouds into textured meshes and render these meshes using Blender. Producing meshes from point clouds is a well-studied, sometimes difficult problem. Point clouds produced by our models often have cracks, outliers, or other types of noise that make the problem particularly challenging. We briefly tried using pre-trained SAP models \citep{sap} for this purpose, but found that the resulting meshes sometimes lost large portions or important details of the shape that were present in the point clouds. Rather than training new SAP models, we opted to take a simpler approach.

To convert point clouds into meshes, we use a regression-based model to predict the signed distance field of an object given its point cloud, and then apply marching cubes \citep{marchingcubes} to the resulting SDF to extract a mesh. We then assign colors to each vertex of the mesh using the color of the nearest point from the original point cloud. For details, see Appendix \ref{app:sdf}.

\section{Results}
\label{sec:results}

In the following sections, we conduct a number of ablations and comparisons to evaluate how our method performs and scales. We adopt the CLIP R-Precision \citep{rprecision} metric for evaluating text-to-3D methods end-to-end, using the same object-centric evaluation prompts as \citet{dreamfields}. Additionally, we introduce a new pair of metrics which we refer to as \textbf{P-IS} and \textbf{P-FID}, which are point cloud analogs for Inception Score \citep{inceptionscore} and FID \citep{fid}, respectively.

To construct our P-IS and P-FID metrics, we employ a modified \mbox{PointNet++} model \citep{pointnet2} to extract features and predict class probabilities for point clouds. For details, see Appendix \ref{app:custompointnet}.

\subsection{Model Scaling and Ablations}
\label{sec:modelablations}

\begin{figure}[t]
    \centering
    \subfigure[P-FID]{
        \includegraphics[width=0.31\textwidth]{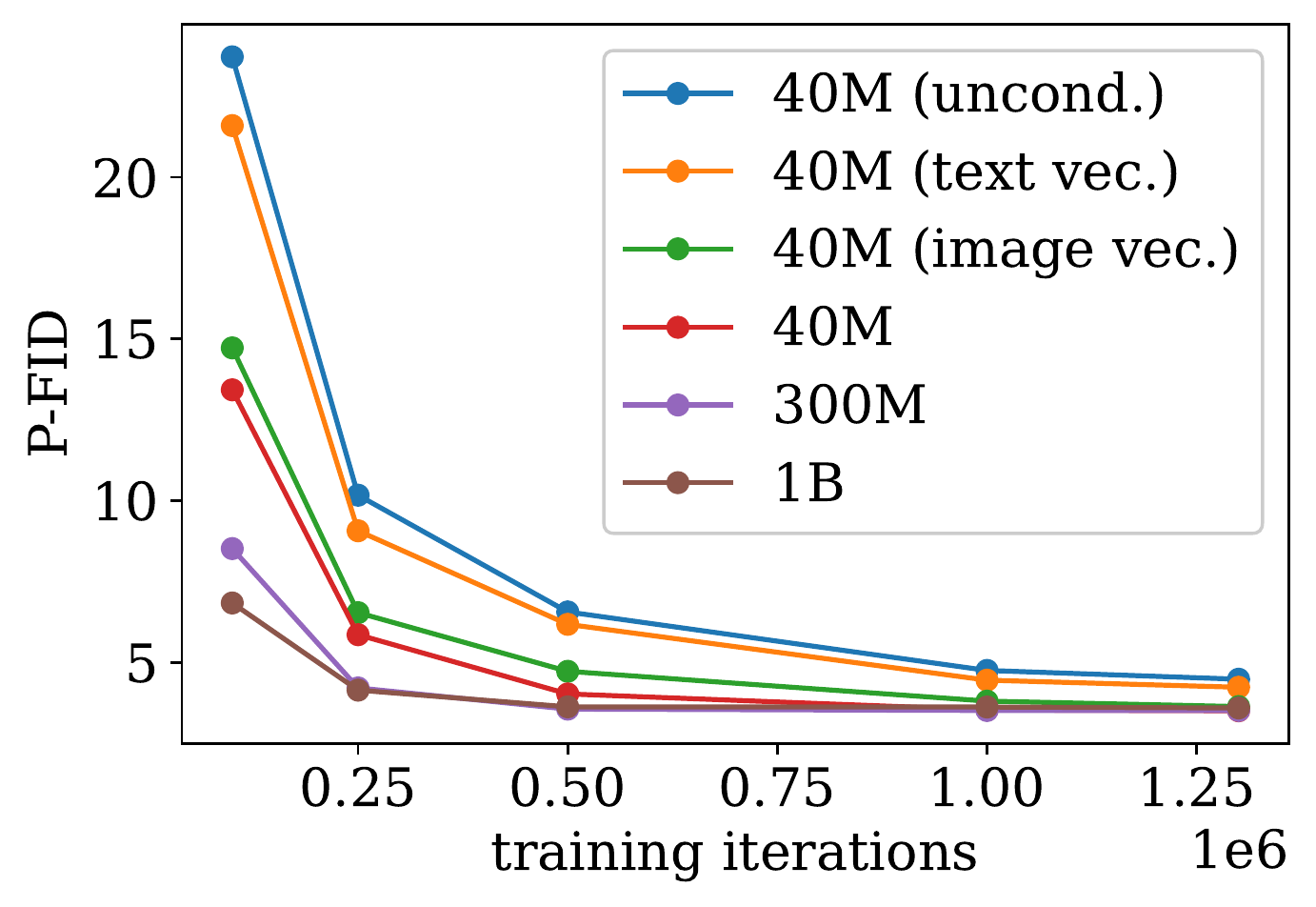}
    }
    \subfigure[P-IS]{
        \includegraphics[width=0.45\textwidth]{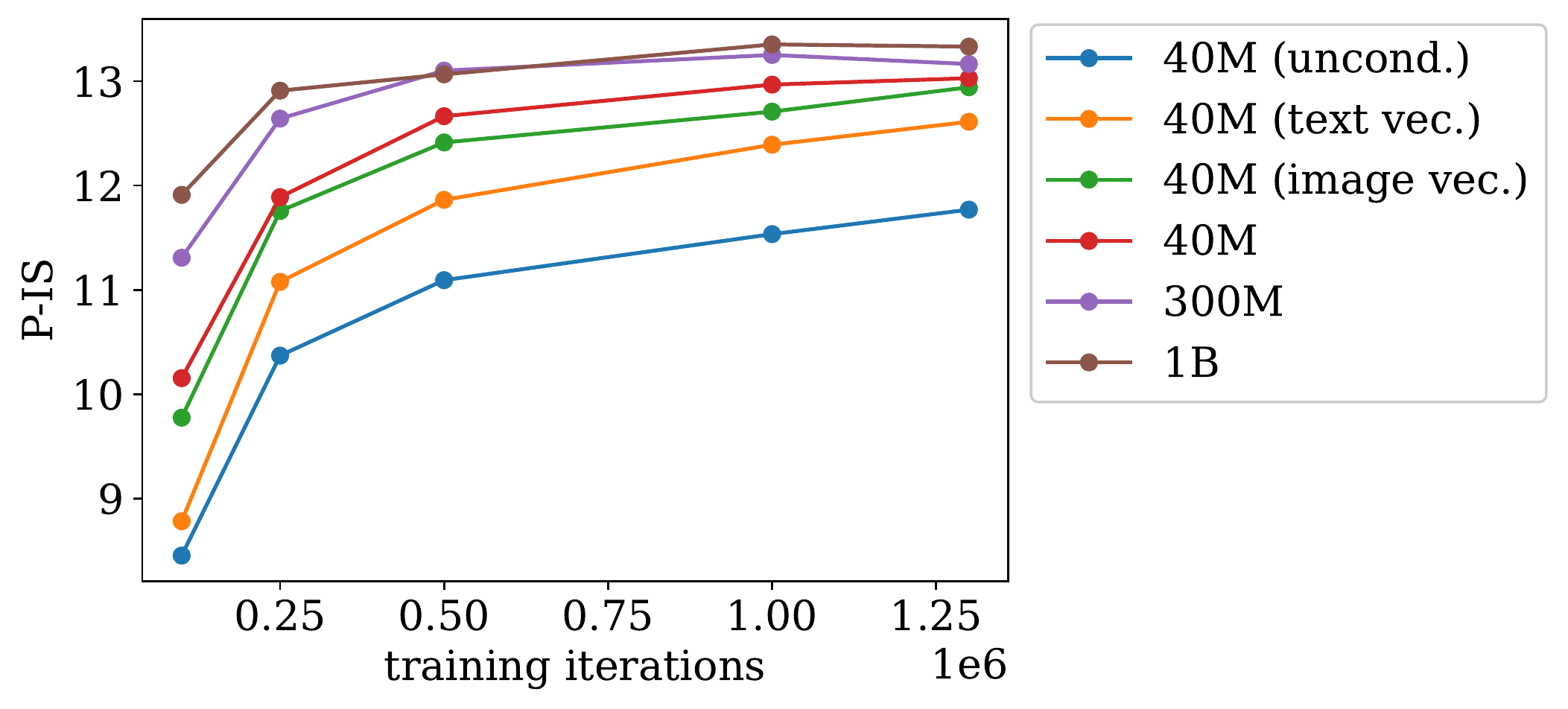}
    }
    \subfigure[CLIP R-Precision]{
        \includegraphics[width=0.45\textwidth]{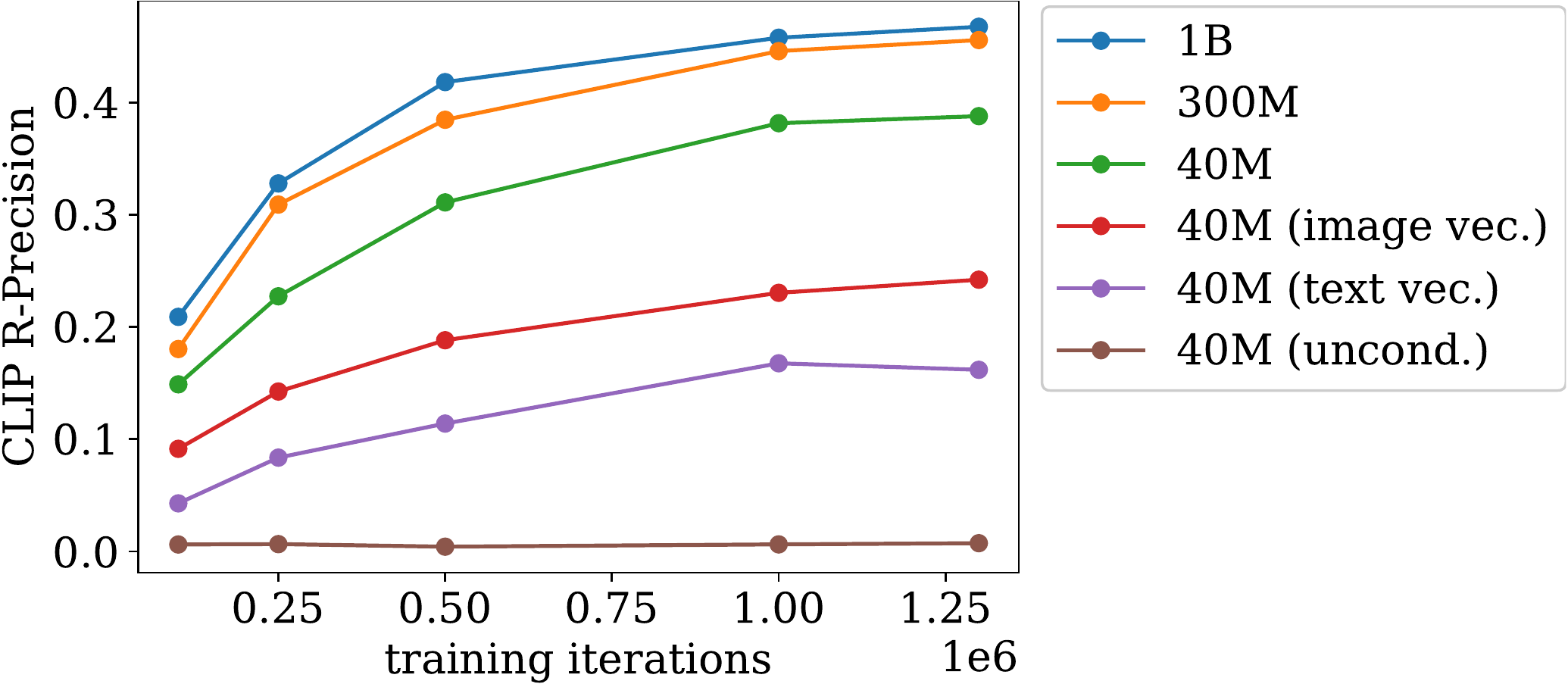}
    }
    \vskip -0.1in
    \caption{\label{fig:train_curves} Sample-based evaluations computed throughout training across different base model runs. The same upsampler and conditioning images are used for all runs.}
\end{figure}

In this section, we train a variety of base diffusion models to study the effect of scaling and to ablate the importance of image conditioning. We train the following base models and evaluate them throughout training:

\begin{itemize}
    \item \textbf{40M (uncond.):} a small model without any conditioning information.
    \item \textbf{40M (text vec.):} a small model which only conditions on text captions, not rendered images. The text caption is embedded with CLIP, and the CLIP embedding is appended as a single extra token of context. This model depends on the text captions present in our 3D dataset, and does not leverage the fine-tuned GLIDE model.
    \item \textbf{40M (image vec.):} a small model which conditions on CLIP image embeddings of rendered images, similar to \citet{clipforge}. This differs from the other image-conditional models in that the image is encoded into a single token of context, rather than as a sequence of latents corresponding to the CLIP latent grid.
    \item \textbf{40M:} a small model with full image conditioning through a grid of CLIP latents.
    \item \textbf{300M:} a medium model with full image conditioning through a grid of CLIP latents.
    \item \textbf{1B:} a large model with full image conditioning through a grid of CLIP latents.
\end{itemize}

In order to isolate changes to the base model, we use the same (image conditional) upsampler model for all evaluations, and use the same 306 pre-generated synthetic views for the CLIP R-Precision evaluation prompts. Here we use the ViT-L/14 CLIP model to compute CLIP R-Precision, but we report results with an alternative CLIP model in \mbox{Section \ref{sec:comparisonother}}.

In Figure \ref{fig:train_curves}, we present the results of our ablations. We find that using only text conditioning with no text-to-image step results in much worse CLIP R-Precision (see Appendix \ref{app:puretext} for more details). Furthermore, we find that using a single CLIP embedding to condition on images is worse than using a grid of embeddings, suggesting that the point  cloud model benefits from seeing more (spatial) information about the conditioning image. Finally, we find that scaling our model improves the speed of P-FID convergence, and increases final CLIP R-Precision.

\subsection{Qualitative Results}
\label{sec:qualitative_results}

We find that \modelname{} can often produce consistent and high-quality 3D shapes for complex prompts. In Figure \ref{fig:banner_images}, we show various point cloud samples which demonstrate our model's ability to infer a variety of shapes while correctly binding colors to the relevant parts of the shapes.

\begin{figure}[t]
    \begin{center}
    \subfigure[Image to point cloud sample for the prompt ``a very realistic 3D rendering of a corgi''.]{
        \includegraphics[width=0.4\textwidth]{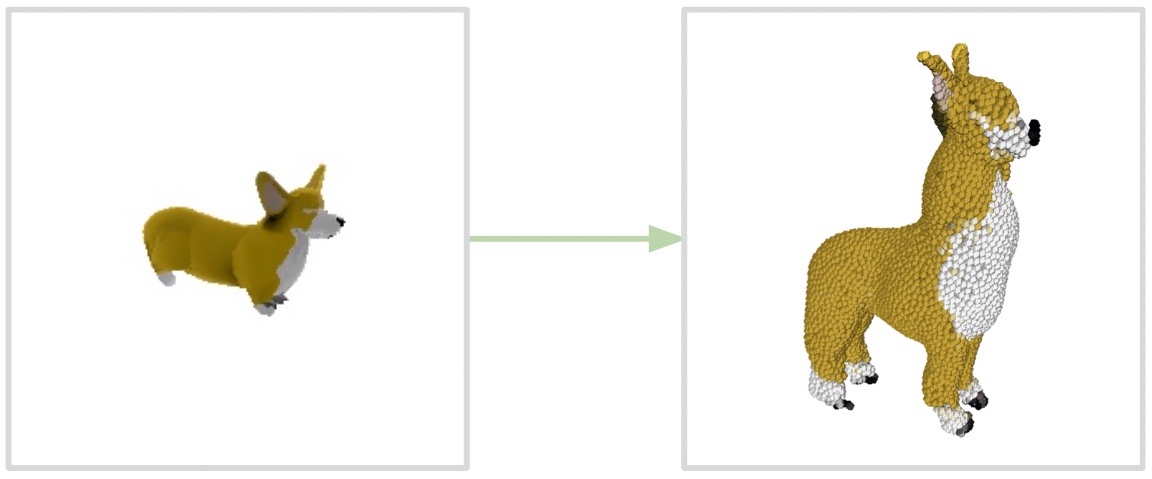}
    }
    \subfigure[Image to point cloud sample for the prompt ``a traffic cone''.]{
        \includegraphics[width=0.4\textwidth]{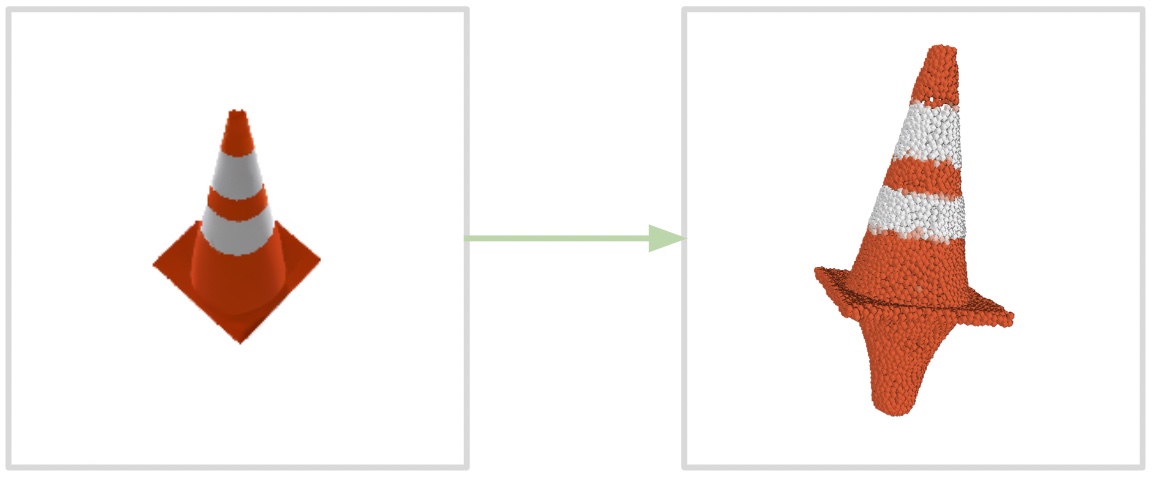}
    }
    \end{center}
    \vskip -0.1in
    \caption{\label{fig:failuremodes} Two common failure modes of our model. In the top example, the model incorrectly interprets the relative proportions of different parts of the depicted object, producing a tall dog instead of a short, long dog. In the bottom example, the model cannot see underneath the traffic cone, and incorrectly infers a second mirrored cone.}
    \vskip -0.1in
\end{figure}

Sometimes the point cloud diffusion model fails to understand or extrapolate the conditioning image, resulting in a shape that does not match the original prompt. We find that this is usually due to one of two issues: 1) the model incorrectly interprets the shape of the object depicted in the image, or 2) the model incorrectly infers some part of the shape that is occluded in the image. In Figure \ref{fig:failuremodes}, we present an example of each of these two failure modes. 

\subsection{Comparison to Other Methods}
\label{sec:comparisonother}

\begin{table}[t]
    \caption{\label{tab:cliprprec} Comparison of \modelname{} to other 3D generative techniques as measured by CLIP R-Precision (with two different CLIP base models) on COCO evaluation prompts. $^*$50 P100-minutes converted to V100-minutes using conversion rate $\frac{1}{3}$. $^{\dagger}$Assuming 2 V100 minutes = 1 A100 minute and 1 TPUv4-minute = 1 A100-minute. We report DreamFields results from \citet{dreamfusion}.}
    \vskip 0.15in
    \centering
    \begin{center}
    \begin{small}
    \begin{tabular}{cccc}
    \toprule
    Method & ViT-B/32 & ViT-L/14 & Latency \\
    \midrule
    DreamFields & 78.6\% & 82.9\% & $\sim 200$ V100-hr$^{\dagger}$ \\
    CLIP-Mesh & 67.8\% & 74.5\% & $\sim 17$ V100-min$^*$ \\
    DreamFusion & 75.1\% & 79.7\% & $\sim 12$ V100-hr$^{\dagger}$ \\
    \midrule
    \makecell{\modelname{} (40M, \\ text-only)} & 15.4\% & 16.2\% & 16 V100-sec \\
    \modelname{} (40M) & 36.5\% & 38.8\% & 1.0 V100-min \\
    \modelname{} (300M) & 40.3\% & 45.6\% & 1.2 V100-min \\
    \modelname{} (1B) & 41.1\% & 46.8\% & 1.5 V100-min \\
    \midrule
    \midrule
    \makecell{Conditioning \\ images} & 69.6\% & 86.6\% & - \\
    \bottomrule
    \end{tabular}
    \end{small}
    \end{center}
    \vskip -0.2in
\end{table}

As text-conditional 3D synthesis is a fairly new area of research, there is not yet a standard set of benchmarks for this task. However, several other works evaluate 3D generation using CLIP R-Precision, and we compare to these methods in Table \ref{tab:cliprprec}. In addition to CLIP R-Precision, we also note the reported sampling compute requirements for each method.

While our method performs worse than the current state-of-the-art, we note two subtleties of this evaluation which may explain some (but likely not all) of this discrepancy:

\begin{itemize}
    \item Unlike multi-view optimization-based methods like DreamFusion, \modelname{} does not explicitly optimize every view to match the text prompt. This could result in lower CLIP R-Precision simply because certain objects are not easy to identify from all angles.
    \item Our method produces point clouds which must be preprocessed before rendering. Converting point clouds into meshes is a difficult problem, and the approach we use can sometimes lose information present in the point clouds themselves.
\end{itemize}

While our method performs worse on this evaluation than state-of-the-art techniques, it produces samples in a small fraction of the time. This could make it more practical for certain applications, or could allow for the discovery of higher-quality 3D objects by sampling many objects and selecting the best one according to some heuristic.

\section{Limitations and Future Work}

While our model is a meaningful step towards fast text-to-3D synthesis, it also has several limitations. Currently, our pipeline requires synthetic renderings, but this limitation could be lifted in the future by training 3D generators that condition on real-world images. Furthermore, while our method produces colored three-dimensional shapes, it does so at a relatively low resolution in a 3D format (point clouds) that does not capture fine-grained shape or texture. Extending this method to produce high-quality 3D representations such as meshes or NeRFs could allow the model's outputs to be used for a variety of applications. Finally, our method could be used to initialize optimization-based techniques to speed up initial convergence.

\begin{figure}[t]
    \centering
    \setlength{\tabcolsep}{2.0pt}
    \begin{tabular}{c}
        \includegraphics[width=0.2\textwidth]{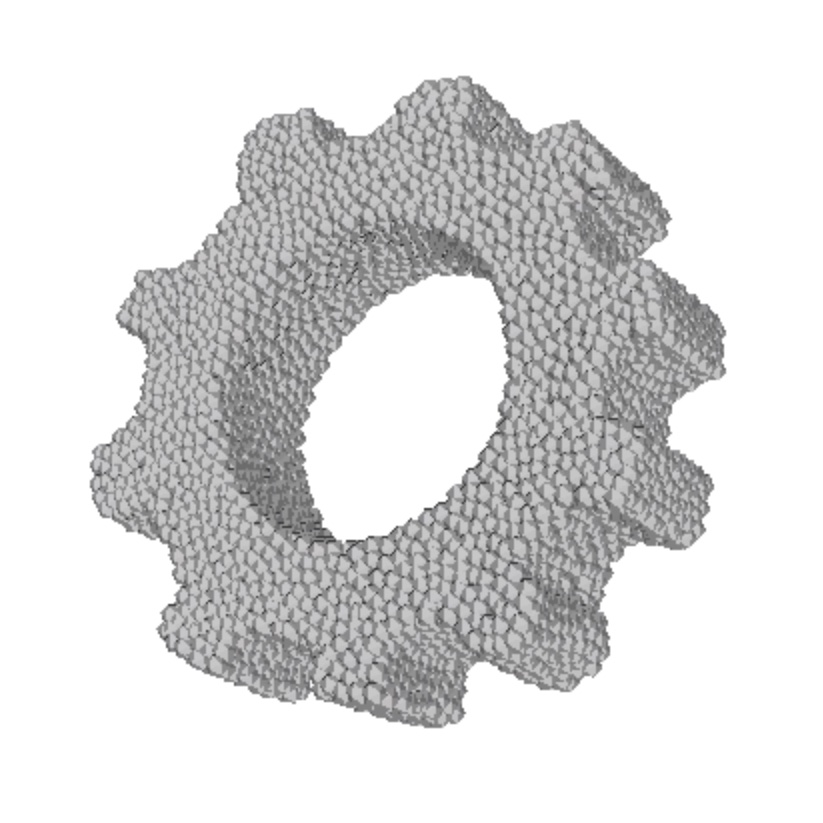} \\

        \scriptsize \makecell{``a 3D printable gear, a single \\ gear 3 inches in diameter \\ and half inch thick''} \\
        \rule{0pt}{0.2pt}
    \end{tabular}

    \caption{Example of a potential misuse of our model, where it could be used to fabricate objects in the real world without external validation.}
    \label{fig:misuseimages}
    \vskip -0.1in
\end{figure}

We expect that this model shares many of the limitations, including bias, as our DALL$\cdot$E 2 system where many of the biases are inherited from the dataset \citep{dalle2card}. In addition, this model has the ability to support the creation of point clouds that can then be used to fabricate products in the real world, for example through 3D printing \citep{3dguns,printingdilemmas,3dkeys}. This has implications both when the models are used to create blueprints for dangerous objects and when the blueprints are trusted to be safe despite no empirical validation (\mbox{Figure \ref{fig:misuseimages}}).

% While our models exhibit some promising capabilities, it is important to recognize that they also inherit many of the risks associated with image generation systems such as . The potential to integrate these models with other technologies, such as 3D printing , could compound these risks even further. Thus, further research and evaluation is necessary in order to better understand and address these concerns.

\section{Conclusion}
\label{sec:conclusion}

We have presented \modelname{}, a system for text-conditional synthesis of 3D point clouds that first generates synthetic views and then generates colored point clouds conditioned on these views. We find that \modelname{} is capable of efficiently producing diverse and complex 3D shapes conditioned on text prompts. We hope that our approach can serve as a starting point for further work in the field of text-to-3D synthesis.

\section{Acknowledgements}
\label{sec:acknowledgements}

We would like to thank everyone behind ChatGPT for creating a tool that helped provide useful writing feedback.

\bibliography{main}

\begin{thebibliography}{79}
\providecommand{\natexlab}[1]{#1}
\providecommand{\url}[1]{\texttt{#1}}
\expandafter\ifx\csname urlstyle\endcsname\relax
  \providecommand{\doi}[1]{doi: #1}\else
  \providecommand{\doi}{doi: \begingroup \urlstyle{rm}\Url}\fi

\bibitem[Achlioptas et~al.(2017)Achlioptas, Diamanti, Mitliagkas, and
  Guibas]{pcautoencoder}
Achlioptas, P., Diamanti, O., Mitliagkas, I., and Guibas, L.
\newblock Learning representations and generative models for 3d point clouds.
\newblock \emph{\href{https://arxiv.org/abs/1707.02392}{arXiv:1707.02392}},
  2017.

\bibitem[Balaji et~al.(2022)Balaji, Nah, Huang, Vahdat, Song, Kreis, Aittala,
  Aila, Laine, Catanzaro, Karras, and Liu]{ediffi}
Balaji, Y., Nah, S., Huang, X., Vahdat, A., Song, J., Kreis, K., Aittala, M.,
  Aila, T., Laine, S., Catanzaro, B., Karras, T., and Liu, M.-Y.
\newblock ediff-i: Text-to-image diffusion models with an ensemble of expert
  denoisers, 2022.

\bibitem[Bautista et~al.(2022)Bautista, Guo, Abnar, Talbott, Toshev, Chen,
  Dinh, Zhai, Goh, Ulbricht, Dehghan, and Susskind]{gaudi}
Bautista, M.~A., Guo, P., Abnar, S., Talbott, W., Toshev, A., Chen, Z., Dinh,
  L., Zhai, S., Goh, H., Ulbricht, D., Dehghan, A., and Susskind, J.
\newblock Gaudi: A neural architect for immersive 3d scene generation.
\newblock \emph{\href{https://arxiv.org/abs/2207.13751}{arXiv:2207.13751}},
  2022.

\bibitem[Cai et~al.(2020)Cai, Yang, Averbuch-Elor, Hao, Belongie, Snavely, and
  Hariharan]{gradientfields}
Cai, R., Yang, G., Averbuch-Elor, H., Hao, Z., Belongie, S., Snavely, N., and
  Hariharan, B.
\newblock Learning gradient fields for shape generation.
\newblock \emph{\href{https://arxiv.org/abs/2008.06520}{arXiv:2008.06520}},
  2020.

\bibitem[Chan et~al.(2020)Chan, Monteiro, Kellnhofer, Wu, and Wetzstein]{pigan}
Chan, E.~R., Monteiro, M., Kellnhofer, P., Wu, J., and Wetzstein, G.
\newblock pi-gan: Periodic implicit generative adversarial networks for
  3d-aware image synthesis.
\newblock \emph{\href{https://arxiv.org/abs/2012.00926}{arXiv:2012.00926}},
  2020.

\bibitem[Chan et~al.(2021)Chan, Lin, Chan, Nagano, Pan, Mello, Gallo, Guibas,
  Tremblay, Khamis, Karras, and Wetzstein]{eg3d}
Chan, E.~R., Lin, C.~Z., Chan, M.~A., Nagano, K., Pan, B., Mello, S.~D., Gallo,
  O., Guibas, L., Tremblay, J., Khamis, S., Karras, T., and Wetzstein, G.
\newblock Efficient geometry-aware 3d generative adversarial networks.
\newblock \emph{\href{https://arxiv.org/abs/2112.07945}{arXiv:2112.07945}},
  2021.

\bibitem[Chen et~al.(2018)Chen, Choy, Savva, Chang, Funkhouser, and
  Savarese]{text2shape}
Chen, K., Choy, C.~B., Savva, M., Chang, A.~X., Funkhouser, T., and Savarese,
  S.
\newblock Text2shape: Generating shapes from natural language by learning joint
  embeddings.
\newblock \emph{\href{https://arxiv.org/abs/1803.08495}{arXiv:1803.08495}},
  2018.

\bibitem[Choy et~al.(2016)Choy, Xu, Gwak, Chen, and Savarese]{r2n2}
Choy, C.~B., Xu, D., Gwak, J., Chen, K., and Savarese, S.
\newblock 3d-r2n2: A unified approach for single and multi-view 3d object
  reconstruction.
\newblock \emph{\href{https://arxiv.org/abs/1604.00449}{arXiv:1604.00449}},
  2016.

\bibitem[Community(2018)]{blender}
Community, B.~O.
\newblock \emph{Blender - a 3D modelling and rendering package}.
\newblock Blender Foundation, Stichting Blender Foundation, Amsterdam, 2018.
\newblock URL \url{http://www.blender.org}.

\bibitem[Dhariwal \& Nichol(2021)Dhariwal and Nichol]{sotapaper}
Dhariwal, P. and Nichol, A.
\newblock Diffusion models beat gans on image synthesis.
\newblock \emph{\href{https://arxiv.org/abs/2105.05233}{arXiv:2105.05233}},
  2021.

\bibitem[Ding et~al.(2021)Ding, Yang, Hong, Zheng, Zhou, Yin, Lin, Zou, Shao,
  Yang, and Tang]{cogview}
Ding, M., Yang, Z., Hong, W., Zheng, W., Zhou, C., Yin, D., Lin, J., Zou, X.,
  Shao, Z., Yang, H., and Tang, J.
\newblock Cogview: Mastering text-to-image generation via transformers.
\newblock \emph{\href{https://arxiv.org/abs/2105.13290}{arXiv:2105.13290}},
  2021.

\bibitem[Fan et~al.(2016)Fan, Su, and Guibas]{pointset}
Fan, H., Su, H., and Guibas, L.
\newblock A point set generation network for 3d object reconstruction from a
  single image.
\newblock \emph{\href{https://arxiv.org/abs/1612.00603}{arXiv:1612.00603}},
  2016.

\bibitem[Feng et~al.(2022)Feng, Zhang, Yu, Fang, Li, Chen, Lu, Liu, Yin, Feng,
  Sun, Tian, Wu, and Wang]{engievilg2}
Feng, Z., Zhang, Z., Yu, X., Fang, Y., Li, L., Chen, X., Lu, Y., Liu, J., Yin,
  W., Feng, S., Sun, Y., Tian, H., Wu, H., and Wang, H.
\newblock Ernie-vilg 2.0: Improving text-to-image diffusion model with
  knowledge-enhanced mixture-of-denoising-experts.
\newblock \emph{\href{https://arxiv.org/abs/2210.15257}{arXiv:2210.15257}},
  2022.

\bibitem[Fu et~al.(2022)Fu, Zhan, Chen, Ritchie, and Sridhar]{shapecrafter}
Fu, R., Zhan, X., Chen, Y., Ritchie, D., and Sridhar, S.
\newblock Shapecrafter: A recursive text-conditioned 3d shape generation model.
\newblock \emph{\href{https://arxiv.org/abs/2207.09446}{arXiv:2207.09446}},
  2022.

\bibitem[Gafni et~al.(2022)Gafni, Polyak, Ashual, Sheynin, Parikh, and
  Taigman]{makeascene}
Gafni, O., Polyak, A., Ashual, O., Sheynin, S., Parikh, D., and Taigman, Y.
\newblock Make-a-scene: Scene-based text-to-image generation with human priors.
\newblock \emph{\href{https://arxiv.org/abs/2203.13131}{arXiv:2203.13131}},
  2022.

\bibitem[Gao et~al.(2022)Gao, Shen, Wang, Chen, Yin, Li, Litany, Gojcic, and
  Fidler]{get3d}
Gao, J., Shen, T., Wang, Z., Chen, W., Yin, K., Li, D., Litany, O., Gojcic, Z.,
  and Fidler, S.
\newblock Get3d: A generative model of high quality 3d textured shapes learned
  from images.
\newblock \emph{\href{https://arxiv.org/abs/2209.11163}{arXiv:2209.11163}},
  2022.

\bibitem[Gkioxari et~al.(2019)Gkioxari, Malik, and Johnson]{meshrcnn}
Gkioxari, G., Malik, J., and Johnson, J.
\newblock Mesh r-cnn.
\newblock \emph{\href{https://arxiv.org/abs/1906.02739}{arXiv:1906.02739}},
  2019.

\bibitem[Goodfellow et~al.(2014)Goodfellow, Pouget-Abadie, Mirza, Xu,
  Warde-Farley, Ozair, Courville, and Bengio]{gan}
Goodfellow, I.~J., Pouget-Abadie, J., Mirza, M., Xu, B., Warde-Farley, D.,
  Ozair, S., Courville, A., and Bengio, Y.
\newblock Generative adversarial networks.
\newblock \emph{\href{https://arxiv.org/abs/1406.2661}{arXiv:1406.2661}}, 2014.

\bibitem[Groueix et~al.(2018)Groueix, Fisher, Kim, Russell, and
  Aubry]{atlasnet}
Groueix, T., Fisher, M., Kim, V.~G., Russell, B.~C., and Aubry, M.
\newblock Atlasnet: A papier-mâché approach to learning 3d surface
  generation.
\newblock \emph{\href{https://arxiv.org/abs/1802.05384}{arXiv:1802.05384}},
  2018.

\bibitem[Gu et~al.(2021)Gu, Liu, Wang, and Theobalt]{stylenerf}
Gu, J., Liu, L., Wang, P., and Theobalt, C.
\newblock Stylenerf: A style-based 3d-aware generator for high-resolution image
  synthesis.
\newblock \emph{\href{https://arxiv.org/abs/2110.08985}{arXiv:2110.08985}},
  2021.

\bibitem[Heusel et~al.(2017)Heusel, Ramsauer, Unterthiner, Nessler, and
  Hochreiter]{fid}
Heusel, M., Ramsauer, H., Unterthiner, T., Nessler, B., and Hochreiter, S.
\newblock Gans trained by a two time-scale update rule converge to a local nash
  equilibrium.
\newblock \emph{Advances in Neural Information Processing Systems 30 (NIPS
  2017)}, 2017.

\bibitem[Ho \& Salimans(2021)Ho and Salimans]{uncond}
Ho, J. and Salimans, T.
\newblock Classifier-free diffusion guidance.
\newblock In \emph{NeurIPS 2021 Workshop on Deep Generative Models and
  Downstream Applications}, 2021.
\newblock URL \url{https://openreview.net/forum?id=qw8AKxfYbI}.

\bibitem[Ho et~al.(2020)Ho, Jain, and Abbeel]{ddpm}
Ho, J., Jain, A., and Abbeel, P.
\newblock Denoising diffusion probabilistic models.
\newblock \emph{\href{https://arxiv.org/abs/2006.11239}{arXiv:2006.11239}},
  2020.

\bibitem[Ho et~al.(2021)Ho, Saharia, Chan, Fleet, Norouzi, and
  Salimans]{cascaded}
Ho, J., Saharia, C., Chan, W., Fleet, D.~J., Norouzi, M., and Salimans, T.
\newblock Cascaded diffusion models for high fidelity image generation.
\newblock \emph{\href{https://arxiv.org/abs/2106.15282}{arXiv:2106.15282}},
  2021.

\bibitem[Ho et~al.(2022{\natexlab{a}})Ho, Chan, Saharia, Whang, Gao, Gritsenko,
  Kingma, Poole, Norouzi, Fleet, and Salimans]{imagenvideo}
Ho, J., Chan, W., Saharia, C., Whang, J., Gao, R., Gritsenko, A., Kingma,
  D.~P., Poole, B., Norouzi, M., Fleet, D.~J., and Salimans, T.
\newblock Imagen video: High definition video generation with diffusion models.
\newblock \emph{\href{https://arxiv.org/abs/2210.02303}{arXiv:2210.02303}},
  2022{\natexlab{a}}.

\bibitem[Ho et~al.(2022{\natexlab{b}})Ho, Salimans, Gritsenko, Chan, Norouzi,
  and Fleet]{videodm}
Ho, J., Salimans, T., Gritsenko, A., Chan, W., Norouzi, M., and Fleet, D.~J.
\newblock Video diffusion models.
\newblock \emph{\href{https://arxiv.org/abs/2204.03458}{arXiv:2204.03458}},
  2022{\natexlab{b}}.

\bibitem[Hong et~al.(2022)Hong, Ding, Zheng, Liu, and Tang]{cogvideo}
Hong, W., Ding, M., Zheng, W., Liu, X., and Tang, J.
\newblock Cogvideo: Large-scale pretraining for text-to-video generation via
  transformers.
\newblock \emph{\href{https://arxiv.org/abs/2205.15868}{arXiv:2205.15868}},
  2022.

\bibitem[Jain et~al.(2021)Jain, Mildenhall, Barron, Abbeel, and
  Poole]{dreamfields}
Jain, A., Mildenhall, B., Barron, J.~T., Abbeel, P., and Poole, B.
\newblock Zero-shot text-guided object generation with dream fields.
\newblock \emph{\href{https://arxiv.org/abs/2112.01455}{arXiv:2112.01455}},
  2021.

\bibitem[Karras et~al.(2022)Karras, Aittala, Aila, and Laine]{edm}
Karras, T., Aittala, M., Aila, T., and Laine, S.
\newblock Elucidating the design space of diffusion-based generative models.
\newblock \emph{\href{https://arxiv.org/abs/2206.00364}{arXiv:2206.00364}},
  2022.

\bibitem[Khalid et~al.(2022)Khalid, Xie, Belilovsky, and Popa]{clipmesh}
Khalid, N.~M., Xie, T., Belilovsky, E., and Popa, T.
\newblock Clip-mesh: Generating textured meshes from text using pretrained
  image-text models.
\newblock \emph{\href{https://arxiv.org/abs/2203.13333}{arXiv:2203.13333}},
  2022.

\bibitem[Kingma \& Welling(2013)Kingma and Welling]{vae}
Kingma, D.~P. and Welling, M.
\newblock Auto-encoding variational bayes.
\newblock \emph{\href{https://arxiv.org/abs/1312.6114}{arXiv:1312.6114}}, 2013.

\bibitem[Kosiorek et~al.(2021)Kosiorek, Strathmann, Zoran, Moreno, Schneider,
  Mokr{\'a}, and Rezende]{nerfvae}
Kosiorek, A.~R., Strathmann, H., Zoran, D., Moreno, P., Schneider, R.,
  Mokr{\'a}, S., and Rezende, D.~J.
\newblock {NeRF-VAE}: A geometry aware {3D} scene generative model.
\newblock \emph{\href{https://arxiv.org/abs/2104.00587}{arXiv:2104.00587}},
  April 2021.

\bibitem[Laine et~al.(2020)Laine, Hellsten, Karras, Seol, Lehtinen, and
  Aila]{nvdiffrast}
Laine, S., Hellsten, J., Karras, T., Seol, Y., Lehtinen, J., and Aila, T.
\newblock Modular primitives for high-performance differentiable rendering.
\newblock \emph{\href{https://arxiv.org/abs/2011.03277}{arXiv:2011.03277}},
  2020.

\bibitem[Lin et~al.(2022{\natexlab{a}})Lin, Gao, Tang, Takikawa, Zeng, Huang,
  Kreis, Fidler, Liu, and Lin]{magic3d}
Lin, C.-H., Gao, J., Tang, L., Takikawa, T., Zeng, X., Huang, X., Kreis, K.,
  Fidler, S., Liu, M.-Y., and Lin, T.-Y.
\newblock Magic3d: High-resolution text-to-3d content creation.
\newblock \emph{\href{https://arxiv.org/abs/2211.10440}{arXiv:2211.10440}},
  2022{\natexlab{a}}.

\bibitem[Lin et~al.(2022{\natexlab{b}})Lin, Yen-Chen, Lai, Lin, Shih, and
  Ramamoorthi]{visionnerf}
Lin, K.-E., Yen-Chen, L., Lai, W.-S., Lin, T.-Y., Shih, Y.-C., and Ramamoorthi,
  R.
\newblock Vision transformer for nerf-based view synthesis from a single input
  image.
\newblock \emph{\href{https://arxiv.org/abs/2207.05736}{arXiv:2207.05736}},
  2022{\natexlab{b}}.

\bibitem[Liu et~al.(2022)Liu, Wang, Qi, and Fu]{towardsimplicit}
Liu, Z., Wang, Y., Qi, X., and Fu, C.-W.
\newblock Towards implicit text-guided 3d shape generation.
\newblock \emph{\href{https://arxiv.org/abs/2203.14622}{arXiv:2203.14622}},
  2022.

\bibitem[Lorensen \& Cline(1987)Lorensen and Cline]{marchingcubes}
Lorensen, W.~E. and Cline, H.~E.
\newblock Marching cubes: A high resolution 3d surface construction algorithm.
\newblock In Stone, M.~C. (ed.), \emph{SIGGRAPH}, pp.\  163--169. ACM, 1987.
\newblock ISBN 0-89791-227-6.
\newblock URL
  \url{http://dblp.uni-trier.de/db/conf/siggraph/siggraph1987.html#LorensenC87}.

\bibitem[Luo \& Hu(2021)Luo and Hu]{flowdiff}
Luo, S. and Hu, W.
\newblock Diffusion probabilistic models for 3d point cloud generation.
\newblock \emph{\href{https://arxiv.org/abs/2103.01458}{arXiv:2103.01458}},
  2021.

\bibitem[Mildenhall et~al.(2020)Mildenhall, Srinivasan, Tancik, Barron,
  Ramamoorthi, and Ng]{nerf}
Mildenhall, B., Srinivasan, P.~P., Tancik, M., Barron, J.~T., Ramamoorthi, R.,
  and Ng, R.
\newblock Nerf: Representing scenes as neural radiance fields for view
  synthesis.
\newblock \emph{\href{https://arxiv.org/abs/2003.08934}{arXiv:2003.08934}},
  2020.

\bibitem[Mishkin et~al.(2022)Mishkin, Ahmad, Brundage, Krueger, and
  Sastry]{dalle2card}
Mishkin, P., Ahmad, L., Brundage, M., Krueger, G., and Sastry, G.
\newblock Dall·e 2 preview - risks and limitations.
\newblock 2022.
\newblock URL
  \url{https://github.com/openai/dalle-2-preview/blob/main/system-card.md}.

\bibitem[Mittal et~al.(2022)Mittal, Cheng, Singh, and Tulsiani]{autosdf}
Mittal, P., Cheng, Y.-C., Singh, M., and Tulsiani, S.
\newblock Autosdf: Shape priors for 3d completion, reconstruction and
  generation.
\newblock \emph{\href{https://arxiv.org/abs/2203.09516}{arXiv:2203.09516}},
  2022.

\bibitem[Mo et~al.(2019)Mo, Guerrero, Yi, Su, Wonka, Mitra, and
  Guibas]{structurenet}
Mo, K., Guerrero, P., Yi, L., Su, H., Wonka, P., Mitra, N., and Guibas, L.~J.
\newblock Structurenet: Hierarchical graph networks for 3d shape generation.
\newblock \emph{\href{https://arxiv.org/abs/1908.00575}{arXiv:1908.00575}},
  2019.

\bibitem[Neely(2016)]{printingdilemmas}
Neely, E.~L.
\newblock The risks of revolution: Ethical dilemmas in 3d printing from a us
  perspective.
\newblock \emph{Science and Engineering Ethics}, 22\penalty0 (5):\penalty0
  1285--1297, Oct 2016.
\newblock ISSN 1471-5546.
\newblock \doi{10.1007/s11948-015-9707-4}.
\newblock URL \url{https://doi.org/10.1007/s11948-015-9707-4}.

\bibitem[Nichol \& Dhariwal(2021)Nichol and Dhariwal]{improved}
Nichol, A. and Dhariwal, P.
\newblock Improved denoising diffusion probabilistic models.
\newblock \emph{\href{https://arxiv.org/abs/2102.09672}{arXiv:2102.09672}},
  2021.

\bibitem[Nichol et~al.(2021)Nichol, Dhariwal, Ramesh, Shyam, Mishkin, McGrew,
  Sutskever, and Chen]{glide}
Nichol, A., Dhariwal, P., Ramesh, A., Shyam, P., Mishkin, P., McGrew, B.,
  Sutskever, I., and Chen, M.
\newblock Glide: Towards photorealistic image generation and editing with
  text-guided diffusion models.
\newblock \emph{\href{https://arxiv.org/abs/2112.10741}{arXiv:2112.10741}},
  2021.

\bibitem[Or-El et~al.(2021)Or-El, Luo, Shan, Shechtman, Park, and
  Kemelmacher-Shlizerman]{stylesdf}
Or-El, R., Luo, X., Shan, M., Shechtman, E., Park, J.~J., and
  Kemelmacher-Shlizerman, I.
\newblock Stylesdf: High-resolution 3d-consistent image and geometry
  generation.
\newblock \emph{\href{https://arxiv.org/abs/2112.11427}{arXiv:2112.11427}},
  2021.

\bibitem[Park et~al.(2021)Park, Azadi, Liu, Darrell, and Rohrbach]{rprecision}
Park, D.~H., Azadi, S., Liu, X., Darrell, T., and Rohrbach, A.
\newblock Benchmark for compositional text-to-image synthesis.
\newblock In \emph{Thirty-fifth Conference on Neural Information Processing
  Systems Datasets and Benchmarks Track (Round 1)}, 2021.
\newblock URL \url{https://openreview.net/forum?id=bKBhQhPeKaF}.

\bibitem[Peng et~al.(2021)Peng, Jiang, Liao, Niemeyer, Pollefeys, and
  Geiger]{sap}
Peng, S., Jiang, C.~M., Liao, Y., Niemeyer, M., Pollefeys, M., and Geiger, A.
\newblock Shape as points: A differentiable poisson solver.
\newblock \emph{\href{https://arxiv.org/abs/2106.03452}{arXiv:2106.03452}},
  2021.

\bibitem[Poole et~al.(2022)Poole, Jain, Barron, and Mildenhall]{dreamfusion}
Poole, B., Jain, A., Barron, J.~T., and Mildenhall, B.
\newblock Dreamfusion: Text-to-3d using 2d diffusion.
\newblock \emph{\href{https://arxiv.org/abs/2209.14988}{arXiv:2209.14988}},
  2022.

\bibitem[Qi et~al.(2017)Qi, Yi, Su, and Guibas]{pointnet2}
Qi, C.~R., Yi, L., Su, H., and Guibas, L.~J.
\newblock Pointnet++: Deep hierarchical feature learning on point sets in a
  metric space.
\newblock \emph{\href{https://arxiv.org/abs/1706.02413}{arXiv:1706.02413}},
  2017.

\bibitem[Radford et~al.(2021)Radford, Kim, Hallacy, Ramesh, Goh, Agarwal,
  Sastry, Askell, Mishkin, Clark, Krueger, and Sutskever]{clip}
Radford, A., Kim, J.~W., Hallacy, C., Ramesh, A., Goh, G., Agarwal, S., Sastry,
  G., Askell, A., Mishkin, P., Clark, J., Krueger, G., and Sutskever, I.
\newblock Learning transferable visual models from natural language
  supervision.
\newblock \emph{\href{https://arxiv.org/abs/2103.00020}{arXiv:2103.00020}},
  2021.

\bibitem[Ramesh et~al.(2021)Ramesh, Pavlov, Goh, Gray, Voss, Radford, Chen, and
  Sutskever]{dalle}
Ramesh, A., Pavlov, M., Goh, G., Gray, S., Voss, C., Radford, A., Chen, M., and
  Sutskever, I.
\newblock Zero-shot text-to-image generation.
\newblock \emph{\href{https://arxiv.org/abs/2102.12092}{arXiv:2102.12092}},
  2021.

\bibitem[Ramesh et~al.(2022)Ramesh, Dhariwal, Nichol, Chu, and Chen]{unclip}
Ramesh, A., Dhariwal, P., Nichol, A., Chu, C., and Chen, M.
\newblock Hierarchical text-conditional image generation with clip latents.
\newblock \emph{\href{https://arxiv.org/abs/2204.06125}{arXiv:2204.06125}},
  2022.

\bibitem[Rombach et~al.(2021)Rombach, Blattmann, Lorenz, Esser, and
  Ommer]{latentdiffusion}
Rombach, R., Blattmann, A., Lorenz, D., Esser, P., and Ommer, B.
\newblock High-resolution image synthesis with latent diffusion models.
\newblock \emph{\href{https://arxiv.org/abs/2112.10752}{arXiv:2112.10752}},
  2021.

\bibitem[Saharia et~al.(2021)Saharia, Ho, Chan, Salimans, Fleet, and
  Norouzi]{sr3}
Saharia, C., Ho, J., Chan, W., Salimans, T., Fleet, D.~J., and Norouzi, M.
\newblock Image super-resolution via iterative refinement.
\newblock
  \emph{\href{https://arxiv.org/abs/arXiv:2104.07636}{arXiv:arXiv:2104.07636}},
  2021.

\bibitem[Saharia et~al.(2022)Saharia, Chan, Saxena, Li, Whang, Denton,
  Ghasemipour, Ayan, Mahdavi, Lopes, Salimans, Ho, Fleet, and Norouzi]{imagen}
Saharia, C., Chan, W., Saxena, S., Li, L., Whang, J., Denton, E., Ghasemipour,
  S. K.~S., Ayan, B.~K., Mahdavi, S.~S., Lopes, R.~G., Salimans, T., Ho, J.,
  Fleet, D.~J., and Norouzi, M.
\newblock Photorealistic text-to-image diffusion models with deep language
  understanding.
\newblock \emph{\href{https://arxiv.org/abs/2205.11487}{arXiv:2205.11487}},
  2022.

\bibitem[Salimans et~al.(2016)Salimans, Goodfellow, Zaremba, Cheung, Radford,
  and Chen]{inceptionscore}
Salimans, T., Goodfellow, I., Zaremba, W., Cheung, V., Radford, A., and Chen,
  X.
\newblock Improved techniques for training gans.
\newblock \emph{\href{https://arxiv.org/abs/1606.03498}{arXiv:1606.03498}},
  2016.

\bibitem[Sanghi et~al.(2021)Sanghi, Chu, Lambourne, Wang, Cheng, Fumero, and
  Malekshan]{clipforge}
Sanghi, A., Chu, H., Lambourne, J.~G., Wang, Y., Cheng, C.-Y., Fumero, M., and
  Malekshan, K.~R.
\newblock Clip-forge: Towards zero-shot text-to-shape generation.
\newblock \emph{\href{https://arxiv.org/abs/2110.02624}{arXiv:2110.02624}},
  2021.

\bibitem[Sanghi et~al.(2022)Sanghi, Fu, Liu, Willis, Shayani, Khasahmadi,
  Sridhar, and Ritchie]{textcraft}
Sanghi, A., Fu, R., Liu, V., Willis, K., Shayani, H., Khasahmadi, A.~H.,
  Sridhar, S., and Ritchie, D.
\newblock Textcraft: Zero-shot generation of high-fidelity and diverse shapes
  from text.
\newblock \emph{\href{https://arxiv.org/abs/2211.01427}{arXiv:2211.01427}},
  2022.

\bibitem[Schwarz et~al.(2020)Schwarz, Liao, Niemeyer, and Geiger]{graf}
Schwarz, K., Liao, Y., Niemeyer, M., and Geiger, A.
\newblock Graf: Generative radiance fields for 3d-aware image synthesis.
\newblock 2020.

\bibitem[Singer et~al.(2022)Singer, Polyak, Hayes, Yin, An, Zhang, Hu, Yang,
  Ashual, Gafni, Parikh, Gupta, and Taigman]{makeavideo}
Singer, U., Polyak, A., Hayes, T., Yin, X., An, J., Zhang, S., Hu, Q., Yang,
  H., Ashual, O., Gafni, O., Parikh, D., Gupta, S., and Taigman, Y.
\newblock Make-a-video: Text-to-video generation without text-video data.
\newblock \emph{\href{https://arxiv.org/abs/2209.14792}{arXiv:2209.14792}},
  2022.

\bibitem[Sohl-Dickstein et~al.(2015)Sohl-Dickstein, Weiss, Maheswaranathan, and
  Ganguli]{dickstein}
Sohl-Dickstein, J., Weiss, E.~A., Maheswaranathan, N., and Ganguli, S.
\newblock Deep unsupervised learning using nonequilibrium thermodynamics.
\newblock \emph{\href{https://arxiv.org/abs/1503.03585}{arXiv:1503.03585}},
  2015.

\bibitem[Song \& Ermon(2020{\natexlab{a}})Song and Ermon]{improvedscore}
Song, Y. and Ermon, S.
\newblock Improved techniques for training score-based generative models.
\newblock \emph{\href{https://arxiv.org/abs/2006.09011}{arXiv:2006.09011}},
  2020{\natexlab{a}}.

\bibitem[Song \& Ermon(2020{\natexlab{b}})Song and Ermon]{scorematching}
Song, Y. and Ermon, S.
\newblock Generative modeling by estimating gradients of the data distribution.
\newblock
  \emph{\href{https://arxiv.org/abs/arXiv:1907.05600}{arXiv:arXiv:1907.05600}},
  2020{\natexlab{b}}.

\bibitem[Song et~al.(2020)Song, Sohl-Dickstein, Kingma, Kumar, Ermon, and
  Poole]{sde}
Song, Y., Sohl-Dickstein, J., Kingma, D.~P., Kumar, A., Ermon, S., and Poole,
  B.
\newblock Score-based generative modeling through stochastic differential
  equations.
\newblock \emph{\href{https://arxiv.org/abs/2011.13456}{arXiv:2011.13456}},
  2020.

\bibitem[Straub \& Kerlin(2016)Straub and Kerlin]{3dkeys}
Straub, J. and Kerlin, S.
\newblock {Evaluation of the use of 3D printing and imaging to create working
  replica keys}.
\newblock In Javidi, B. and Son, J.-Y. (eds.), \emph{Three-Dimensional Imaging,
  Visualization, and Display 2016}, volume 9867, pp.\  98670E. International
  Society for Optics and Photonics, SPIE, 2016.
\newblock \doi{10.1117/12.2223858}.
\newblock URL \url{https://doi.org/10.1117/12.2223858}.

\bibitem[Sun et~al.(2018)Sun, Huh, Liao, Zhang, and Lim]{multiviewnovel}
Sun, S.-H., Huh, M., Liao, Y.-H., Zhang, N., and Lim, J.~J.
\newblock Multi-view to novel view: Synthesizing novel views with self-learned
  confidence.
\newblock In Ferrari, V., Hebert, M., Sminchisescu, C., and Weiss, Y. (eds.),
  \emph{Computer Vision -- ECCV 2018}, pp.\  162--178, Cham, 2018. Springer
  International Publishing.
\newblock ISBN 978-3-030-01219-9.

\bibitem[van~den Oord et~al.(2017)van~den Oord, Vinyals, and
  Kavukcuoglu]{vqvae}
van~den Oord, A., Vinyals, O., and Kavukcuoglu, K.
\newblock Neural discrete representation learning.
\newblock \emph{\href{https://arxiv.org/abs/1711.00937}{arXiv:1711.00937}},
  2017.

\bibitem[Vaswani et~al.(2017)Vaswani, Shazeer, Parmar, Uszkoreit, Jones, Gomez,
  Kaiser, and Polosukhin]{transformer}
Vaswani, A., Shazeer, N., Parmar, N., Uszkoreit, J., Jones, L., Gomez, A.~N.,
  Kaiser, L., and Polosukhin, I.
\newblock Attention is all you need.
\newblock \emph{\href{https://arxiv.org/abs/1706.03762}{arXiv:1706.03762}},
  2017.

\bibitem[Walther(2014)]{3dguns}
Walther, G.
\newblock Printing insecurity? the security implications of 3d-printing of
  weapons.
\newblock \emph{Science and engineering ethics}, 21, 12 2014.
\newblock \doi{10.1007/s11948-014-9617-x}.

\bibitem[Wang et~al.(2018)Wang, Zhang, Li, Fu, Liu, and Jiang]{pixel2mesh}
Wang, N., Zhang, Y., Li, Z., Fu, Y., Liu, W., and Jiang, Y.-G.
\newblock Pixel2mesh: Generating 3d mesh models from single rgb images.
\newblock \emph{\href{https://arxiv.org/abs/1804.01654}{arXiv:1804.01654}},
  2018.

\bibitem[Watson et~al.(2022)Watson, Chan, Martin-Brualla, Ho, Tagliasacchi, and
  Norouzi]{3dim}
Watson, D., Chan, W., Martin-Brualla, R., Ho, J., Tagliasacchi, A., and
  Norouzi, M.
\newblock Novel view synthesis with diffusion models.
\newblock \emph{\href{https://arxiv.org/abs/2210.04628}{arXiv:2210.04628}},
  2022.

\bibitem[Wu et~al.(2015)Wu, Song, Khosla, Yu, Zhang, Tang, and Xiao]{modelnet}
Wu, Z., Song, S., Khosla, A., Yu, F., Zhang, L., Tang, X., and Xiao, J.
\newblock 3d shapenets: A deep representation for volumetric shapes.
\newblock In \emph{Proceedings of the IEEE Conference on Computer Vision and
  Pattern Recognition (CVPR)}, June 2015.

\bibitem[Yang et~al.(2019)Yang, Huang, Hao, Liu, Belongie, and
  Hariharan]{pointflow}
Yang, G., Huang, X., Hao, Z., Liu, M.-Y., Belongie, S., and Hariharan, B.
\newblock Pointflow: 3d point cloud generation with continuous normalizing
  flows.
\newblock \emph{\href{https://arxiv.org/abs/1906.12320}{arXiv:1906.12320}},
  2019.

\bibitem[Yu et~al.(2020)Yu, Ye, Tancik, and Kanazawa]{pixelnerf}
Yu, A., Ye, V., Tancik, M., and Kanazawa, A.
\newblock pixelnerf: Neural radiance fields from one or few images.
\newblock \emph{\href{https://arxiv.org/abs/2012.02190}{arXiv:2012.02190}},
  2020.

\bibitem[Yu et~al.(2022)Yu, Xu, Koh, Luong, Baid, Wang, Vasudevan, Ku, Yang,
  Ayan, Hutchinson, Han, Parekh, Li, Zhang, Baldridge, and Wu]{parti}
Yu, J., Xu, Y., Koh, J.~Y., Luong, T., Baid, G., Wang, Z., Vasudevan, V., Ku,
  A., Yang, Y., Ayan, B.~K., Hutchinson, B., Han, W., Parekh, Z., Li, X.,
  Zhang, H., Baldridge, J., and Wu, Y.
\newblock Scaling autoregressive models for content-rich text-to-image
  generation.
\newblock \emph{\href{https://arxiv.org/abs/2206.10789}{arXiv:2206.10789}},
  2022.

\bibitem[Zeng et~al.(2022)Zeng, Vahdat, Williams, Gojcic, Litany, Fidler, and
  Kreis]{lion}
Zeng, X., Vahdat, A., Williams, F., Gojcic, Z., Litany, O., Fidler, S., and
  Kreis, K.
\newblock Lion: Latent point diffusion models for 3d shape generation.
\newblock \emph{\href{https://arxiv.org/abs/2210.06978}{arXiv:2210.06978}},
  2022.

\bibitem[Zhou et~al.(2021{\natexlab{a}})Zhou, Du, and Wu]{pvd}
Zhou, L., Du, Y., and Wu, J.
\newblock 3d shape generation and completion through point-voxel diffusion.
\newblock \emph{\href{https://arxiv.org/abs/2104.03670}{arXiv:2104.03670}},
  2021{\natexlab{a}}.

\bibitem[Zhou et~al.(2021{\natexlab{b}})Zhou, Xie, Ni, and Tian]{cips3d}
Zhou, P., Xie, L., Ni, B., and Tian, Q.
\newblock Cips-3d: A 3d-aware generator of gans based on
  conditionally-independent pixel synthesis.
\newblock \emph{\href{https://arxiv.org/abs/2110.09788}{arXiv:2110.09788}},
  2021{\natexlab{b}}.

\end{thebibliography}
\bibliographystyle{icml2022}

\clearpage

\appendix

\section{Hyperparameters}

\begin{table}[t]
    \caption{\label{tab:trainhyperparams} Training hyper-parameters for our point cloud diffusion models. Width and depth refer to the size of the transformer backbone.}
    \vskip 0.15in
    \centering
    \begin{center}
    \begin{small}
    \begin{tabular}{ccccc}
    \toprule
    Model & Width & Depth & LR & \# Params \\
    \midrule
    Base (40M) & 512 & 12 & 1e-4 & 40,466,956 \\
    Base (300M) & 1024 & 24 & 7e-5 & 311,778,316 \\
    Base (1B) & 2048 & 24 & 5e-5 & 1,244,311,564 \\
    Upsampler & 512 & 12 & 1e-4 & 40,470,540 \\
    \bottomrule
    \end{tabular}
    \end{small}
    \end{center}
\end{table}

We train all of our diffusion models with batch size 64 for 1,300,000 iterations. In Table \ref{tab:trainhyperparams}, we enumerate the training hyperparameters that were varied across model sizes. We train all of our models with 1024 diffusion timesteps. For our upsampler model, we use the linear noise schedule from \citet{ddpm}, and for our base models, we use the cosine noise schedule proposed by \citet{improved}.

\begin{table}[t]
    \caption{\label{tab:samplehyperparams} Sampling hyperparameters for figures and CLIP R-Precision evaluations.}
    \vskip 0.15in
    \centering
    \begin{center}
    \begin{small}
    \begin{tabular}{ccc}
    \toprule
    Hyperparameter & Base & Upsampler \\
    \midrule
    Timesteps & 64 & 64 \\
    Guidance scale & 3.0 & 3.0 \\
    $S_{\text{churn}}$ & 3 & 0 \\
    $\sigma_{\text{min}}$ & 1e-3 & 1e-3 \\
    $\sigma_{\text{max}}$ & 120 & 160 \\
    \bottomrule
    \end{tabular}
    \end{small}
    \end{center}
\end{table}

For P-FID and P-IS evaluations, we produce 10K samples using stochastic DDPM with the full noise schedule. For CLIP R-Precision and figures in the paper, we use 64 steps (128 function evaluations) of the Heun sampler from \citet{edm} for both the base and upsampler models. Table \ref{tab:samplehyperparams} enumerates the hyperparameters used for Heun sampling. When sampling from GLIDE, we use 150 diffusion steps for the base model, and 50 diffusion steps for the upsampling model. We report sampling time for each component of our stack in Table \ref{tab:sampletime}.

\begin{table}[t]
    \caption{\label{tab:sampletime} Sampling performance for various components of our model. We use the Karras sampler for our base and upsampler models, but not for GLIDE.}
    \vskip 0.15in
    \centering
    \begin{center}
    \begin{small}
    \begin{tabular}{ccc}
    \toprule
    Model & V100 seconds \\
    \midrule
    GLIDE & 46.28 \\
    Upsampler (40M) & 12.58 \\
    Base (40M) & 3.35 \\
    Base (300M) & 12.78 \\
    Base (1B) & 28.67 \\
    \bottomrule
    \end{tabular}
    \end{small}
    \end{center}
\end{table}

\section{P-FID and P-IS Metrics}
\label{app:custompointnet}

To evaluate P-FID and P-IS, we train a \mbox{PointNet++} model on ModelNet40 \citep{modelnet} using an open source implementation.\footnote{\url{https://github.com/yanx27/Pointnet_Pointnet2_pytorch}} We modify the baseline model in several ways. First, we double the width of the model, resulting in roughly 16 million parameters. Next, we apply some additional data augmentations to make the model more robust to out-of-distribution samples. In particular, we apply random rotations to each point cloud, and we add Gaussian noise to the points with standard deviation sampled from $U[0, 0.01]$.

To compute P-FID, we extract features for each point cloud from the layer before the final ReLU activation. To compute P-IS, we use the predicted class probabilities for the 40 classes from ModelNet40. We note that our generative models are trained on a dataset which only has P-IS 12.95, so our best reported P-IS score of $\sim 13$ is near the expected upper bound.

\section{Mesh Extraction}
\label{app:sdf}

To convert point clouds into meshes, we train a model which predicts SDFs from point clouds and apply marching cubes to the resulting SDFs. We parametrize our SDF model as an encoder-decoder Transformer. First, an 8-layer encoder processes the input point cloud as an unordered sequence, producing a sequence of hidden representations. Then, a 4-layer cross-attention decoder takes 3D coordinates and the sequence of latent vectors, and predicts an SDF value. Each input query point is processed independently, allowing for efficient batching. Using more layers in the encoder and fewer in the decoder allows us to amortize the encoding cost across many query points.

We train our SDF regression model on a subset of 2.4 million manifold meshes from our dataset, and add Gaussian noise with $\sigma=0.005$ to the point clouds as data augmentation. We train the model $f_{\theta}(x)$ to predict the SDF $y$ with a weighted L1 objective:

\[ \begin{cases} 
      1 \cdot ||f_\theta(x) - y||_1 & f_\theta(x) > y \\
      4 \cdot ||f_\theta(x) - y||_1 & f_\theta(x) \le y
   \end{cases}
\]

Here, we define the SDF such that points outside of the surface have negative sign. Therefore, in the face of uncertainty, the model is encouraged predict that points are inside the surface. We found this to be helpful in initial experiments, likely because it helps prevent the resulting meshes from effectively ignoring thin or noisy parts of the point cloud.

\begin{figure}[t]
    \centering
    \setlength{\tabcolsep}{2.0pt}
    \begin{tabular}{cc}
        \includegraphics[width=0.2\textwidth]{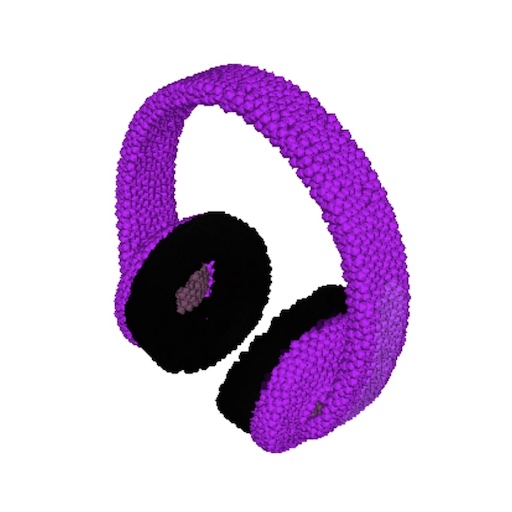} & \includegraphics[width=0.2\textwidth]{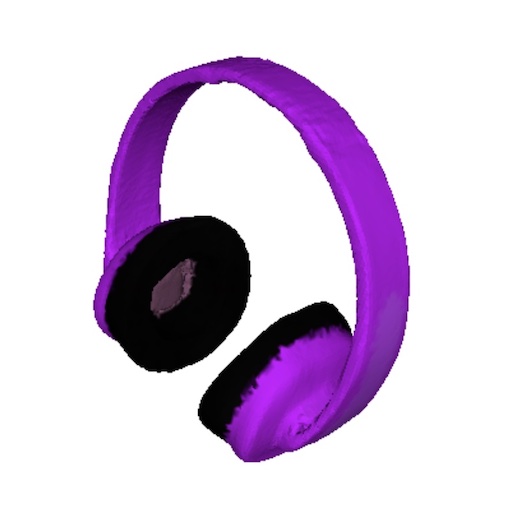} \\
        \includegraphics[width=0.2\textwidth]{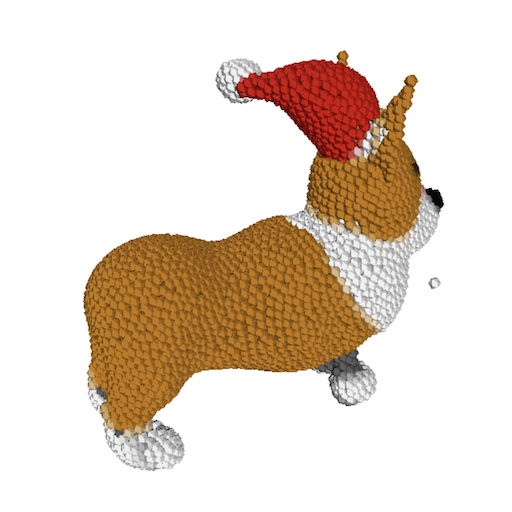} & \includegraphics[width=0.2\textwidth]{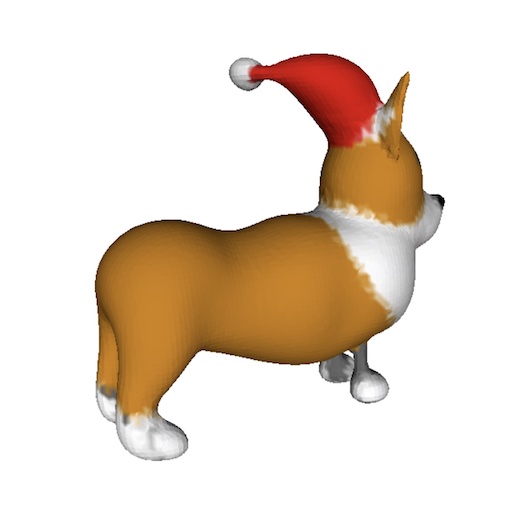} \\
        \includegraphics[width=0.2\textwidth]{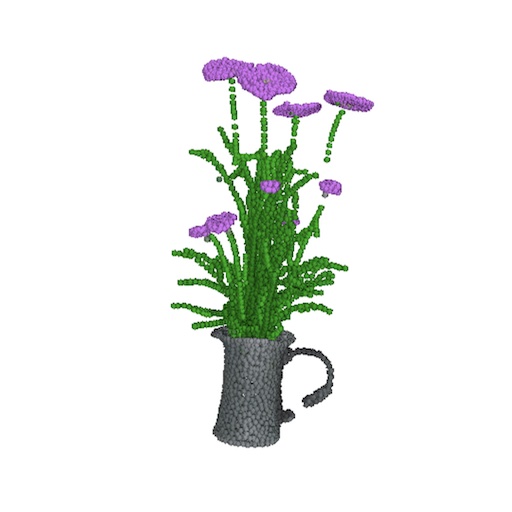} & \includegraphics[width=0.2\textwidth]{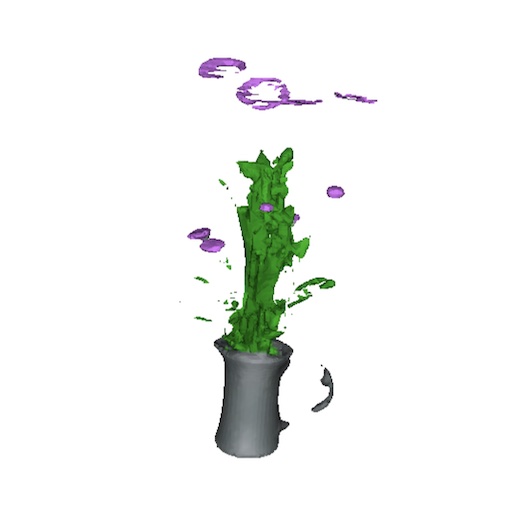}
    \end{tabular}

    \caption{Examples of point clouds (left) and corresponding extracted meshes (right). We find that our method often produces smooth meshes and removes outliers (middle row), but can sometimes miss thin/sparse parts of objects (bottom row).}
    \label{fig:meshrecon}
    \vskip -0.1in
\end{figure}

When producing meshes for evaluations, we use a grid size of $128 \times 128 \times 128$, resulting in $128^3$ queries to the SDF model. In Figure \ref{fig:meshrecon}, we show examples of input point clouds and corresponding output meshes from our model. We observe that our method works well in many cases, but sometimes fails to capture thin or sparse parts of a point cloud.

\section{Conditioning on \texorpdfstring{DALL$\cdot$E}{DALL-E} 2 Samples}
\label{app:dalle2renders}

\begin{figure}[t]
    \centering
    \setlength{\tabcolsep}{2.0pt}
    \begin{tabular}{cc}
        \includegraphics[width=0.2\textwidth]{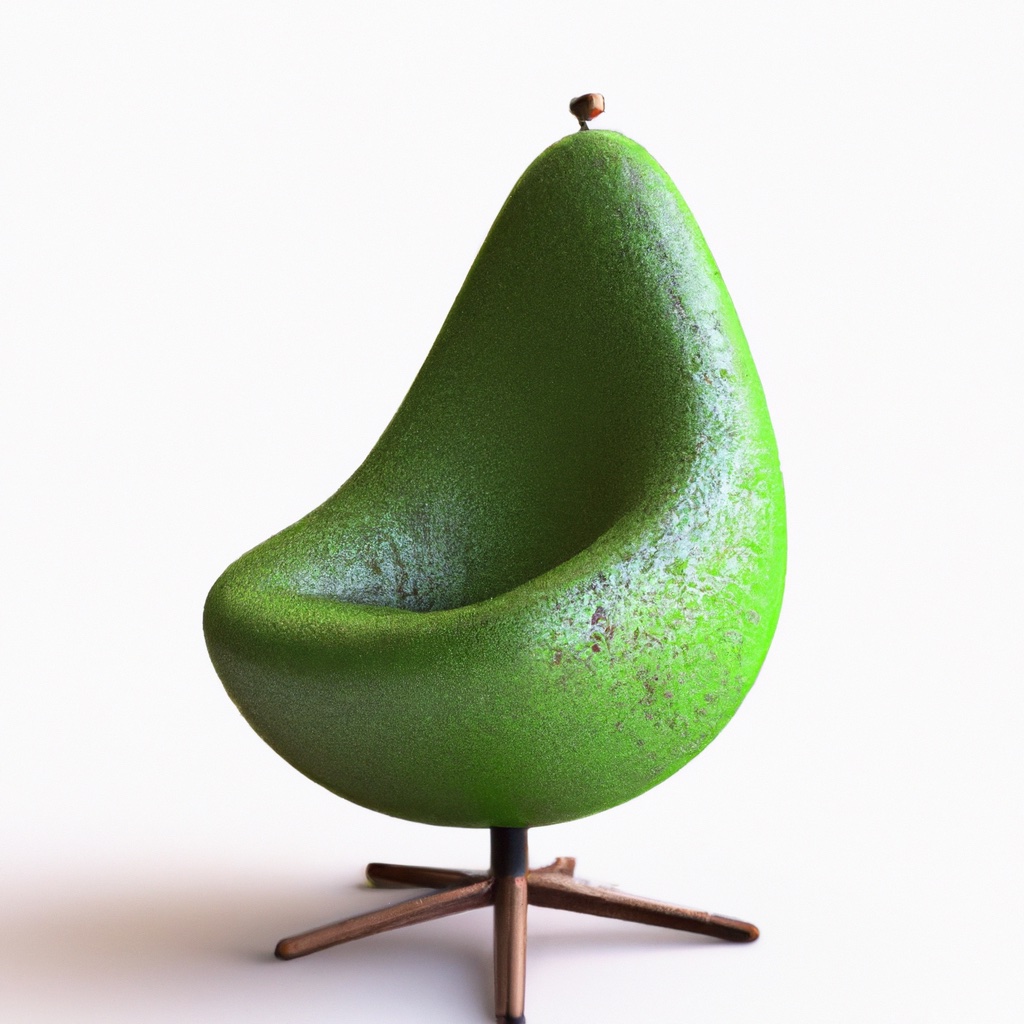} & \includegraphics[width=0.2\textwidth]{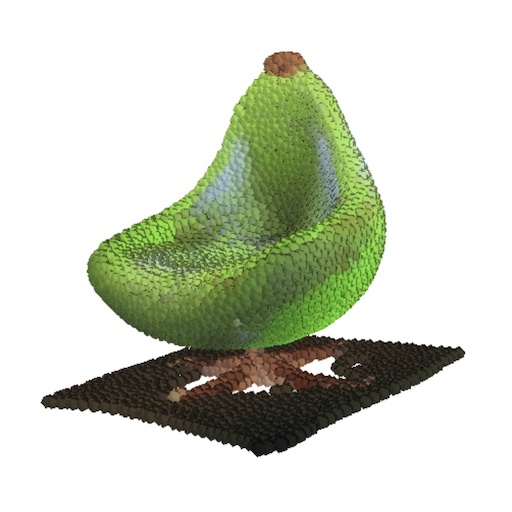} \\
        \includegraphics[width=0.2\textwidth]{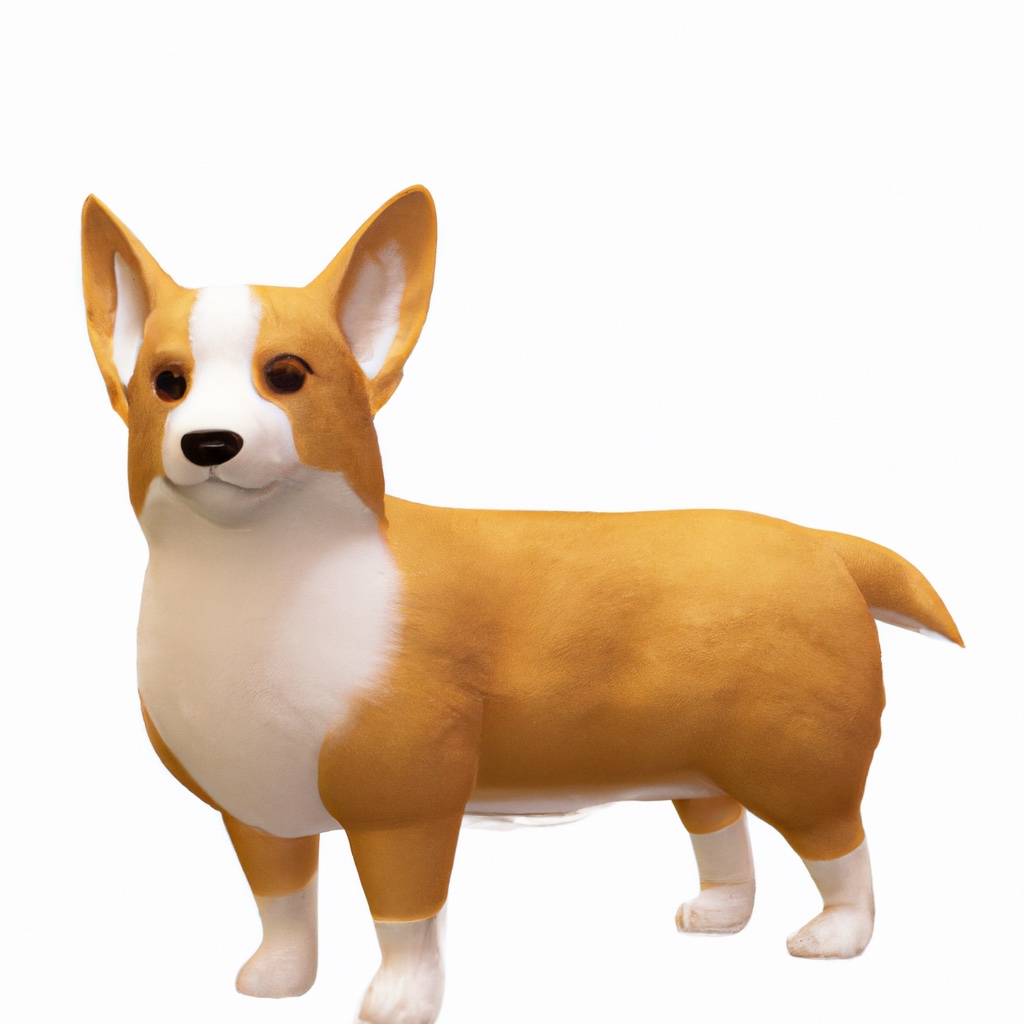} & \includegraphics[width=0.2\textwidth]{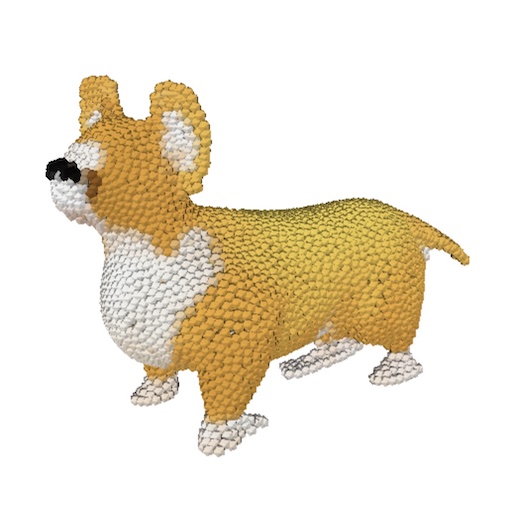} \\
        \includegraphics[width=0.2\textwidth]{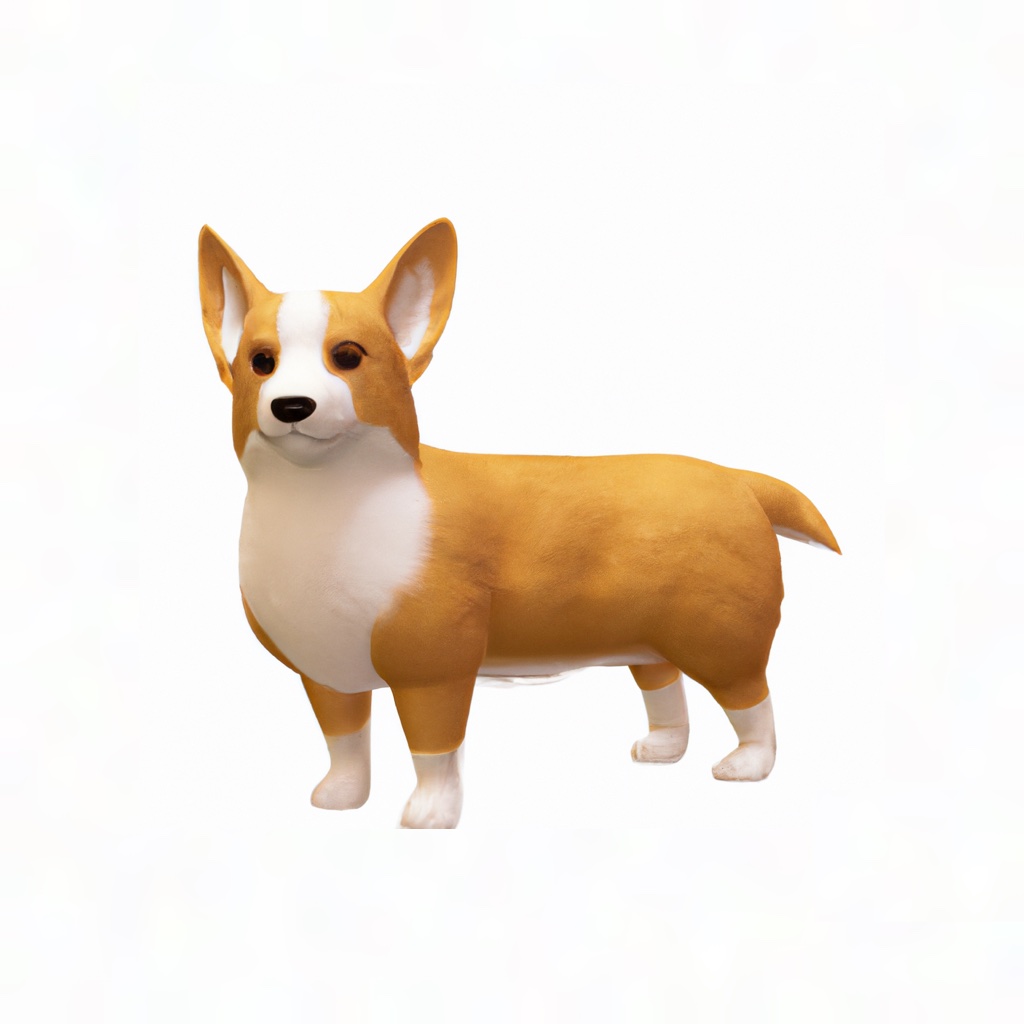} & \includegraphics[width=0.2\textwidth]{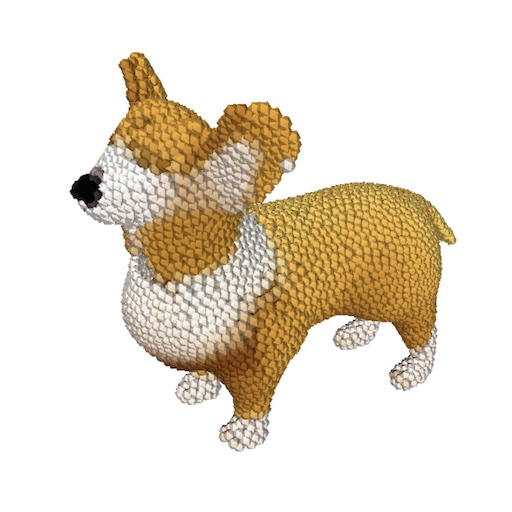}
    \end{tabular}

    \caption{Examples of point clouds reconstructed from DALL$\cdot$2 generations. The top image was produced using the prompt ``a 3d rendering of an avocado chair, chair imitating an avocado, full view, white background''. The middle image was produced using the prompt ``a simple full view of a 3d rendering of a corgi in front of a white background''. The bottom image is the same as the middle image, but with an additional white border.}
    \label{fig:dalle2recon}
    \vskip -0.1in
\end{figure}

In our main experiments, we use a specialized text-to-image model to produce in-distribution conditioning images for our point cloud models. In this section, we explore what happens if we use renders from a pre-existing text-to-image model, DALL$\cdot$E 2.

In Figure \ref{fig:dalle2recon}, we present three image-to-3D examples where the conditioning images are generated by DALL$\cdot$E 2. We find that DALL$\cdot$E 2 tends to include shadows under objects, and our point cloud model interprets these as a dark ground plane. We also find that our point cloud model can misinterpret shapes from the generated images when the objects take up too much of the image. In these cases, adding a border around the generated images can improve the reconstructed shapes.

\section{Pure Text-Conditional Generation}
\label{app:puretext}

In Section \ref{sec:modelablations}, we train a pure text-conditional point cloud model without an additional image generation step. While we find that this model performs worse on evaluations than our full system, it still achieves non-trivial results. In this section, we explore the capabilities and limitations of this model.

\begin{figure}[t]
    \centering
    \setlength{\tabcolsep}{2.0pt}
    \begin{tabular}{cc}
        \includegraphics[width=0.2\textwidth]{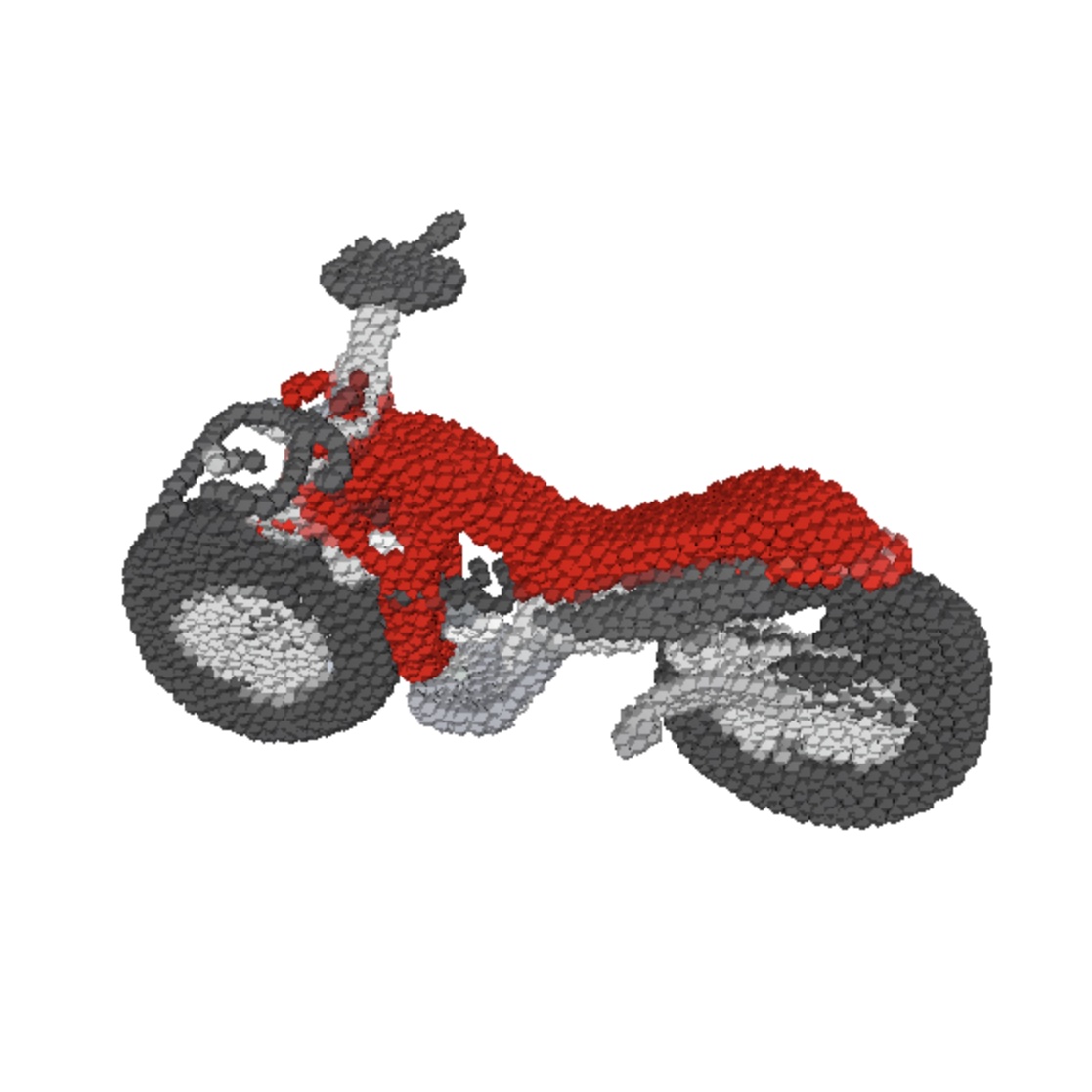} &
        \includegraphics[width=0.2\textwidth]{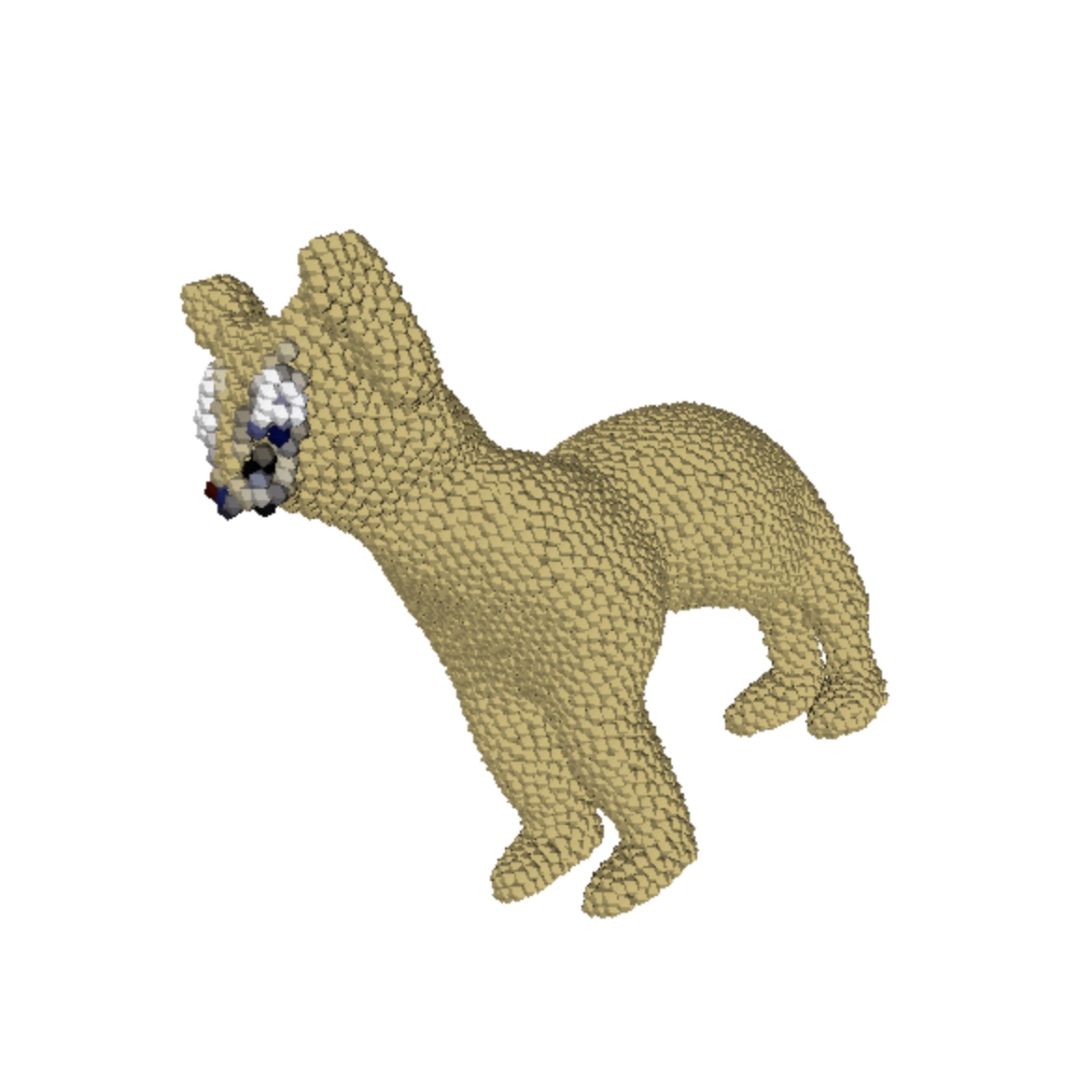} \\

        \scriptsize \makecell{``a motorbike''} &
        \scriptsize \makecell{``a dog''} \\
        \rule{0pt}{0.2pt} \\

        \includegraphics[width=0.2\textwidth]{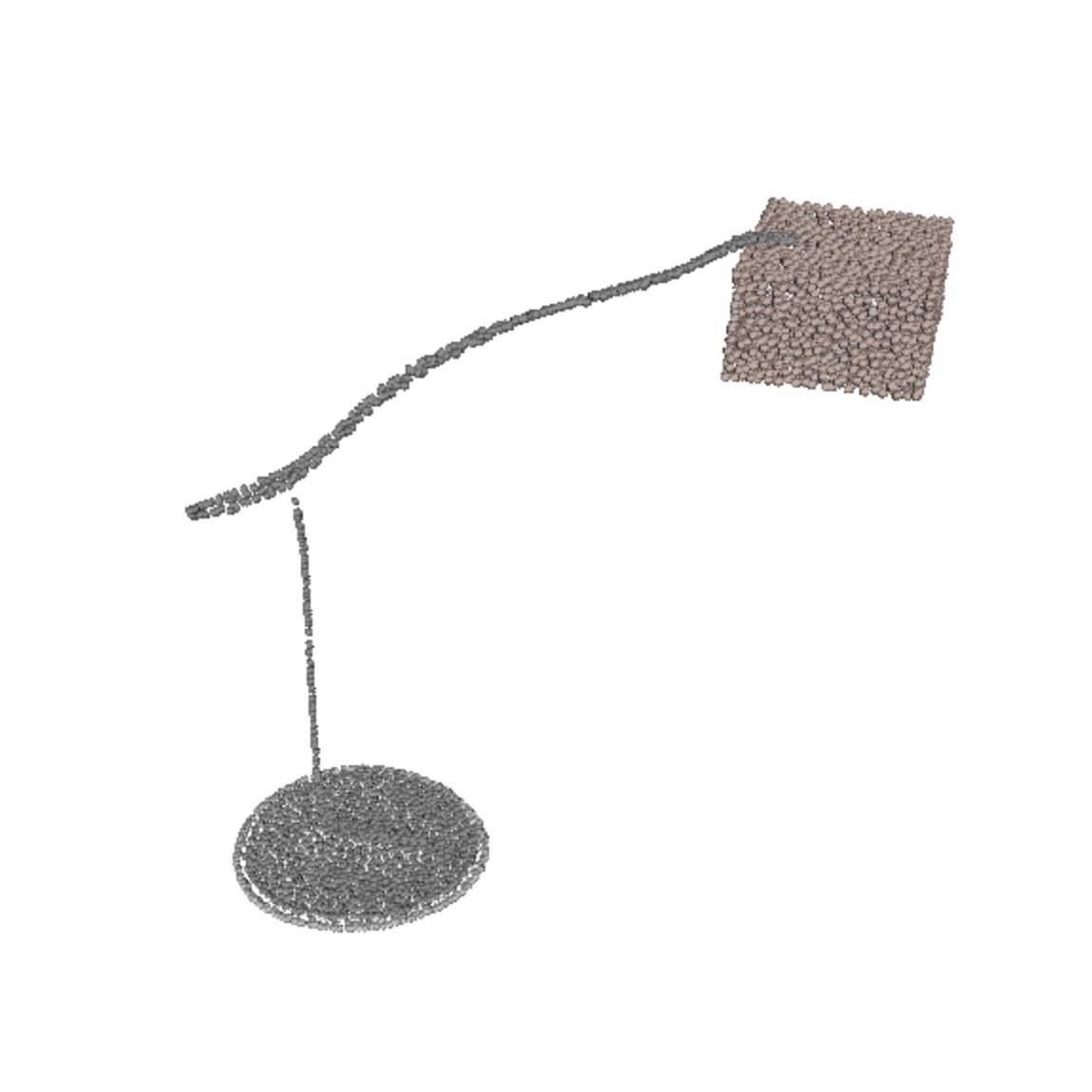} &
        \includegraphics[width=0.2\textwidth]{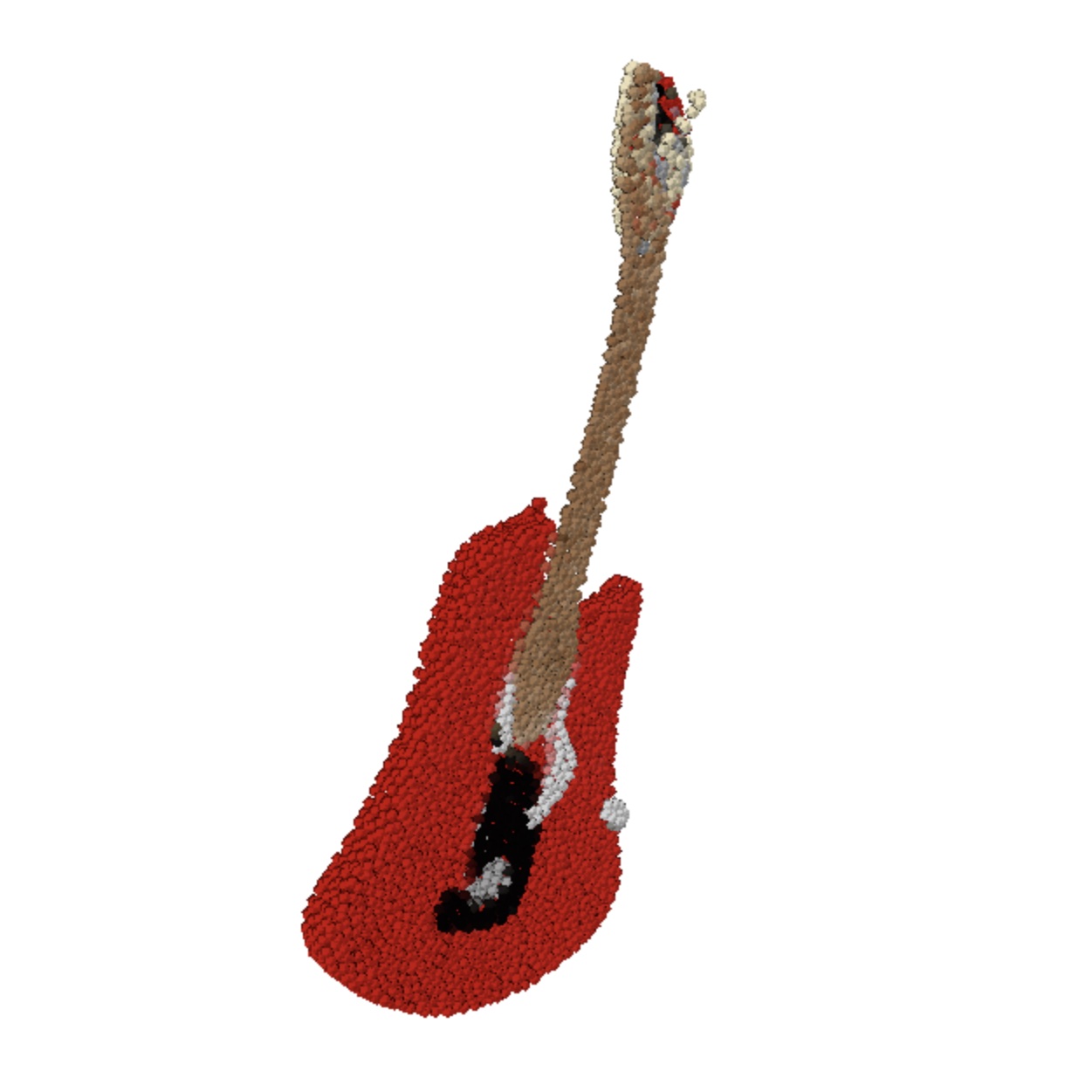} \\

        \scriptsize \makecell{``a desk lamp''} &
        \scriptsize \makecell{``a guitar''} \\
        \rule{0pt}{0.2pt} \\

        \includegraphics[width=0.2\textwidth]{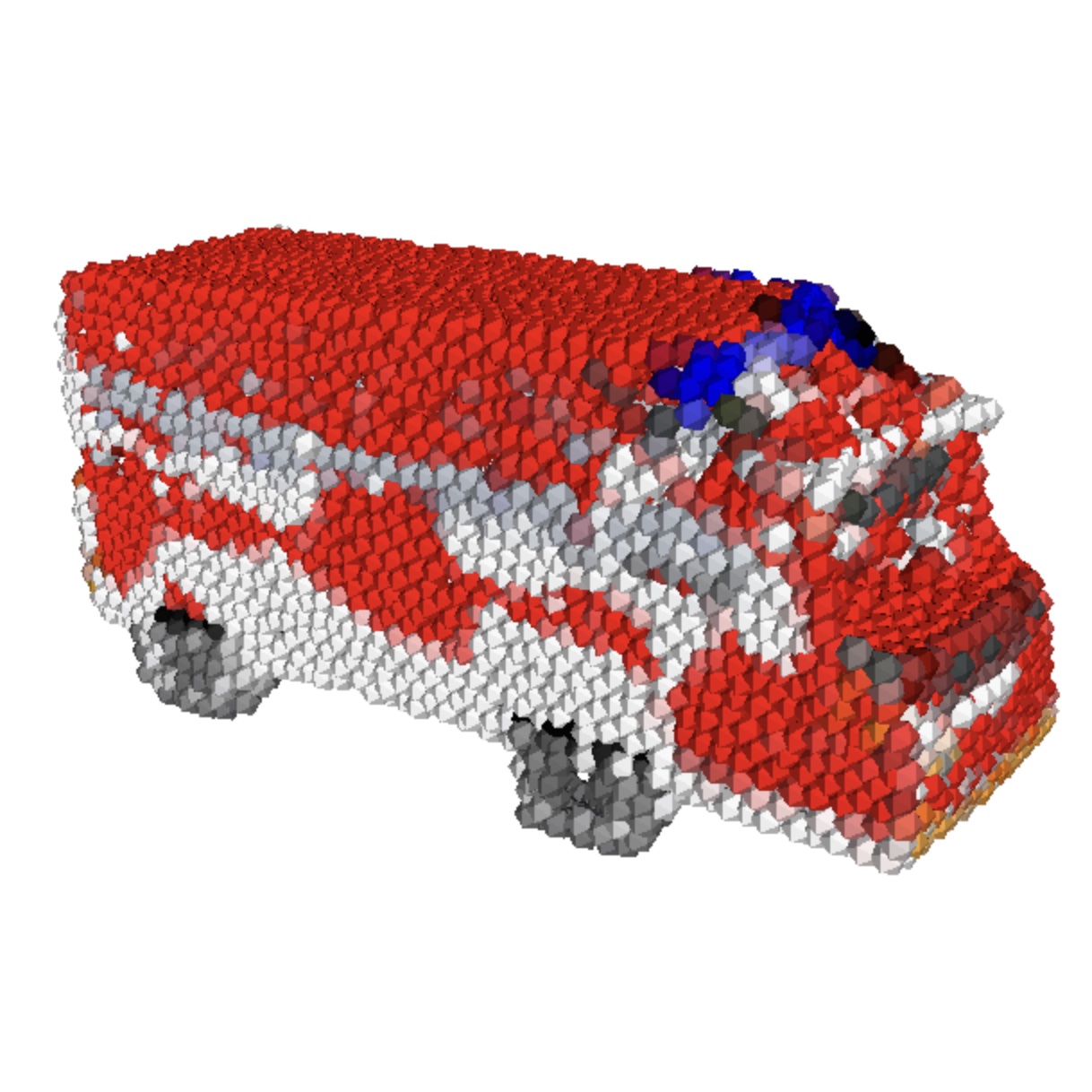} &
        \includegraphics[width=0.2\textwidth]{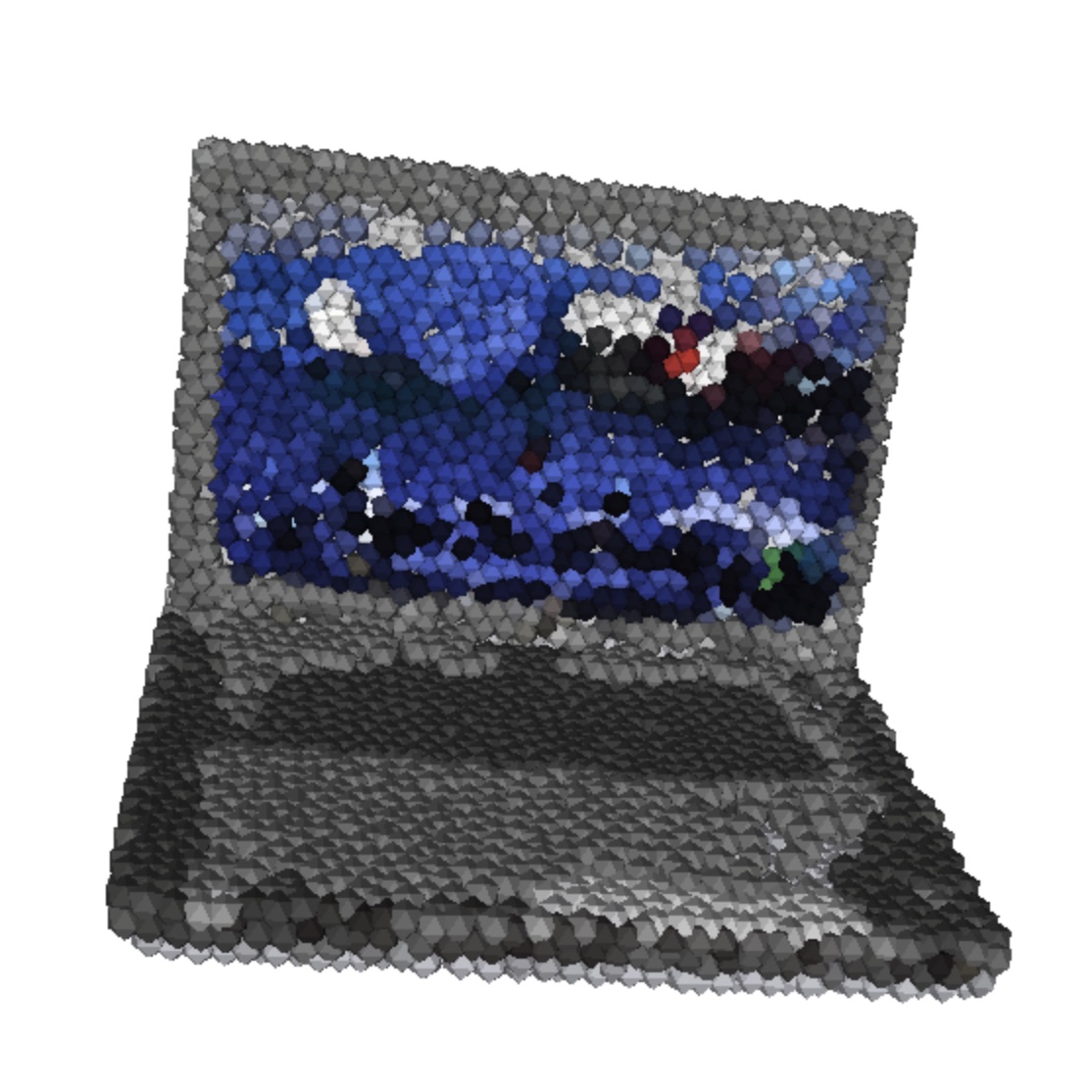} \\

        \scriptsize \makecell{``an ambulance''} &
        \scriptsize \makecell{``a laptop computer''} \\
        \rule{0pt}{0.2pt}
    \end{tabular}

    \caption{Selected point clouds generated by our pure text-conditional 40M parameter point cloud diffusion model.}
    \label{fig:textcondimages}
    \vskip -0.1in
\end{figure}

\begin{figure}[h]
    \centering
    \subfigure[Prompt: ``a small red cube is sitting on top of a large blue cube. red on top, blue on bottom'']{
        \includegraphics[width=0.45\textwidth]{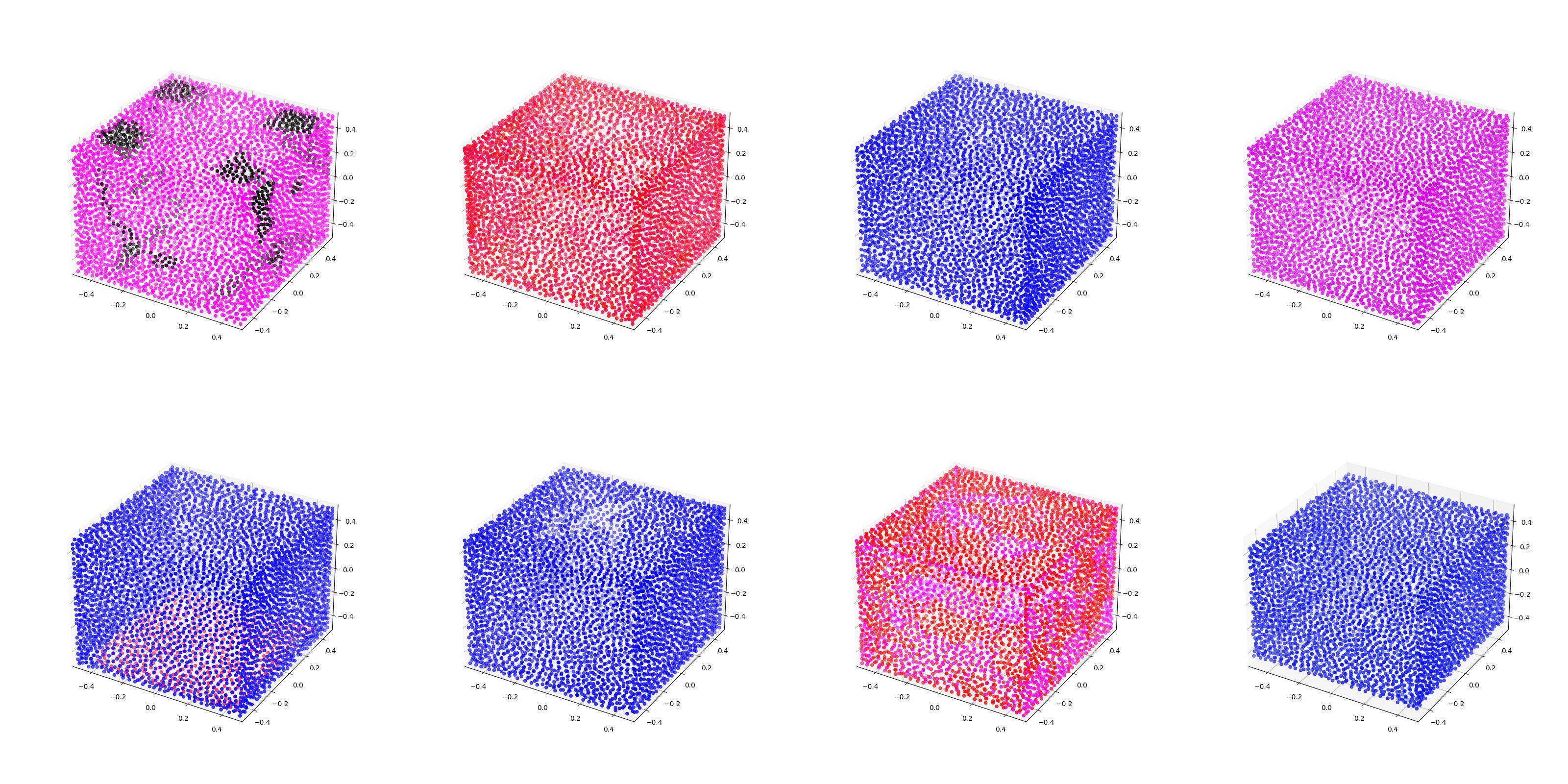}
    }
    \subfigure[Prompt: ``a corgi wearing a red santa hat'']{
        \includegraphics[width=0.45\textwidth]{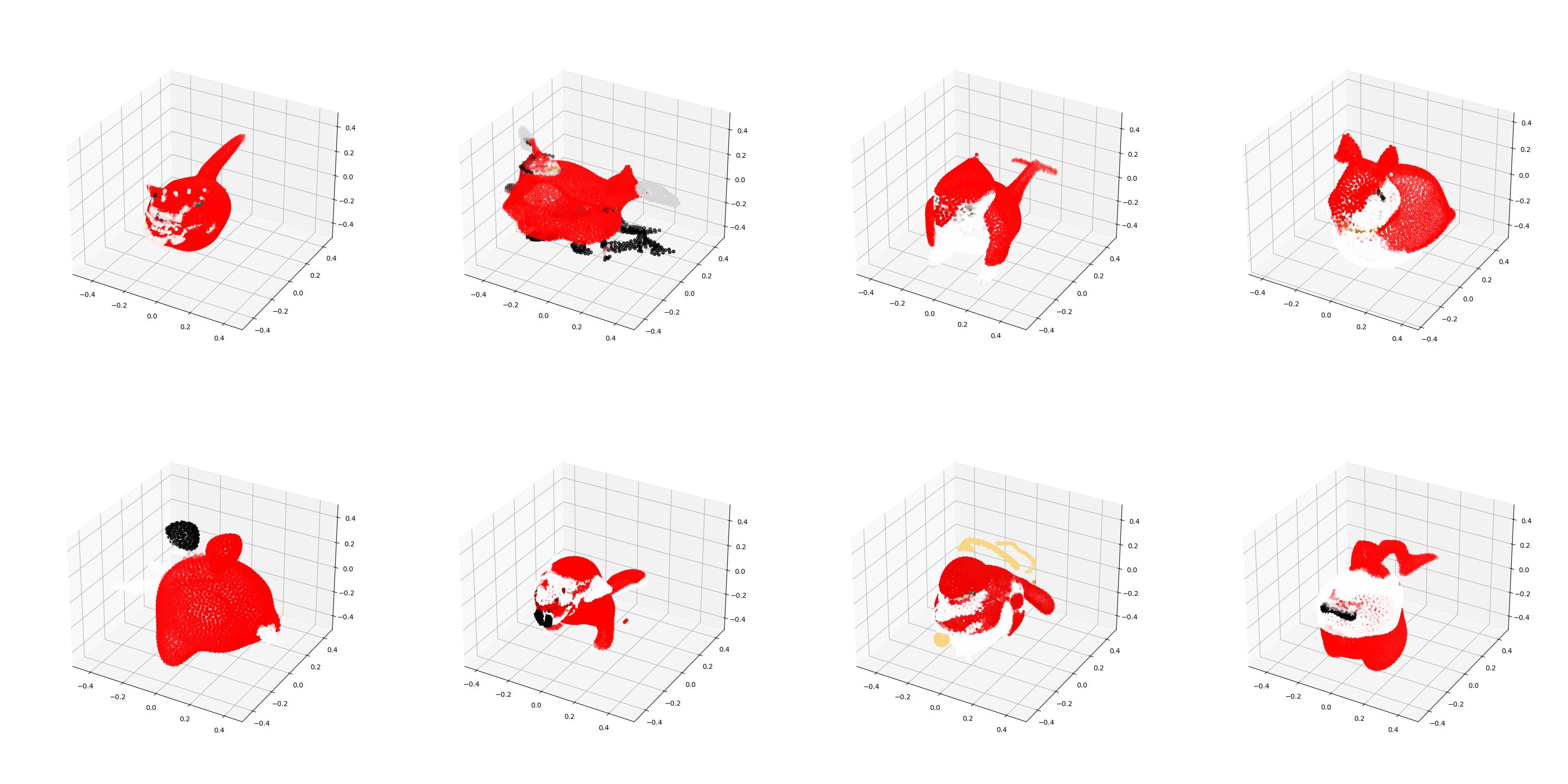}
    }
    \vskip -0.1in
    \caption{\label{fig:textcondfailures} Sample grids where our small, pure text-conditional model fails to understand complex prompts.}
\end{figure}

In Figure \ref{fig:textcondimages}, we show examples where our text-conditional model is able to produce point clouds matching the provided text prompt. Notably, these examples include simple prompts that describe single objects. In Figure \ref{fig:textcondfailures}, we show examples where this model struggles with prompts that combine multiple concepts.

\begin{figure}[h]
    \centering
    \subfigure[Prompt: ``a man'']{
        \includegraphics[width=0.45\textwidth]{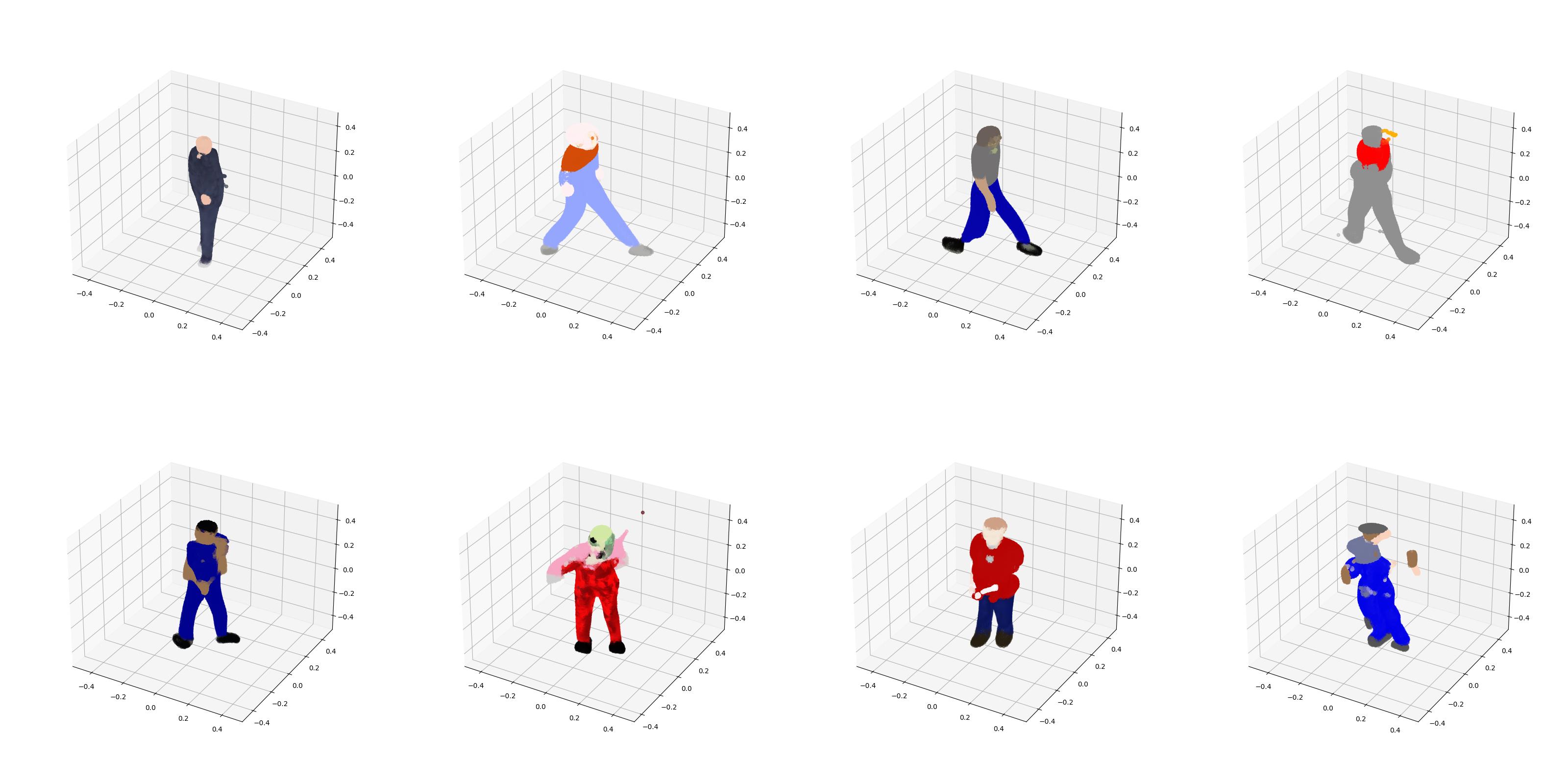}
    }
    \subfigure[Prompt: ``a woman'']{
        \includegraphics[width=0.45\textwidth]{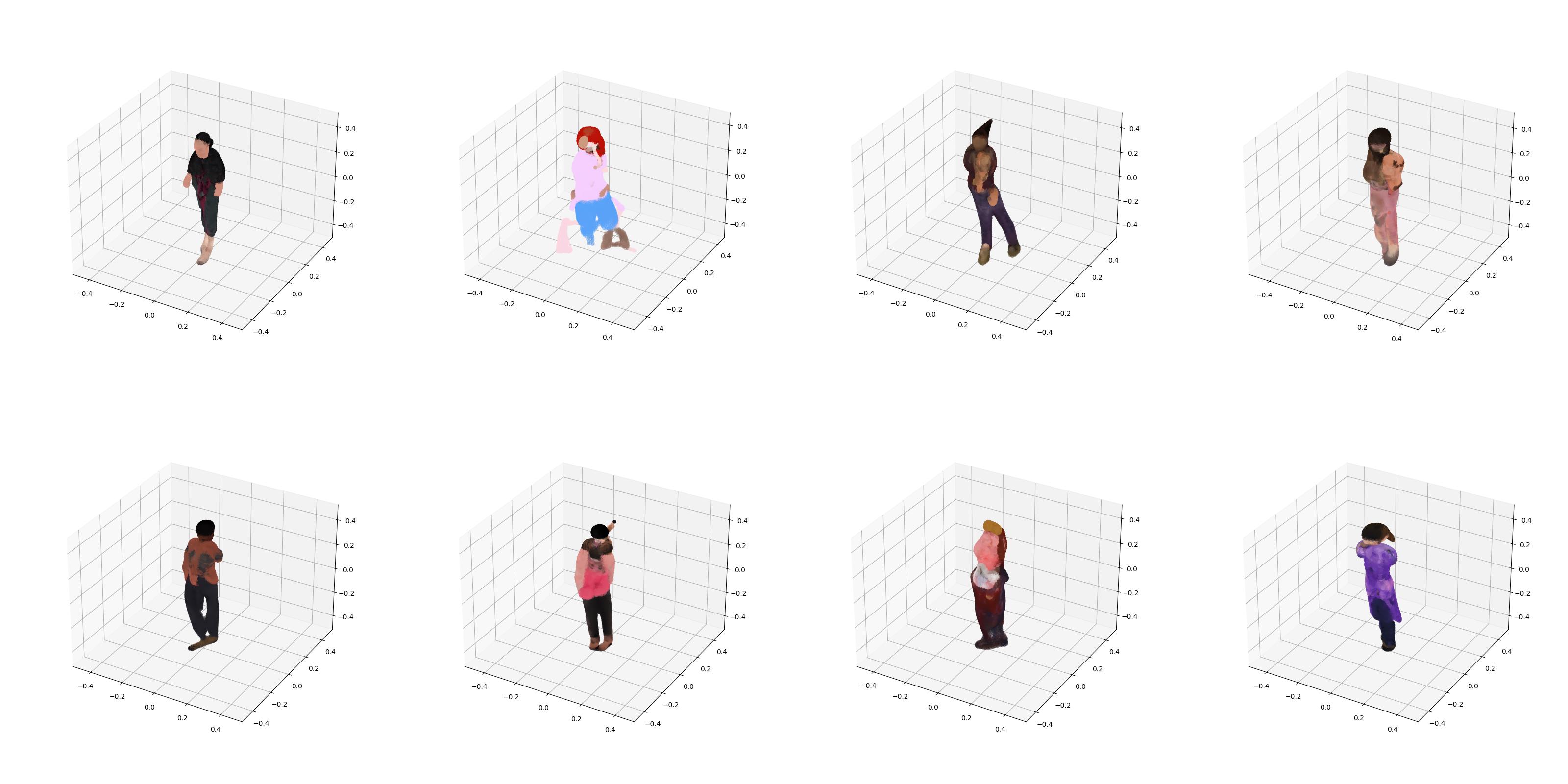}
    }
    \vskip -0.1in
    \caption{\label{fig:biasexamples} Sample grids from our pure text-conditional 40M parameter model. Samples in the top grid use the same noise seed as the corresponding samples in the bottom grid.}
\end{figure}

Finally, we expect that this model has inherited biases from our 3D dataset. We present one possible example of this in Figure \ref{fig:biasexamples}, wherein the model produces longer and narrower objects for the prompt ``a woman'' than for the prompt ``a man'' when using a fixed diffusion noise seed.

% \begin{figure}[h]
%     \centering
%     \setlength{\tabcolsep}{2.0pt}
%     \begin{tabular}{cc}
%         \includegraphics[width=0.2\textwidth]{figures/text_cond/a red cube on top of a blue cube.jpg} &
%         \includegraphics[width=0.2\textwidth]{figures/text_cond/a corgi wearing a santa hat.jpg} \\

%         \scriptsize \makecell{``a red cube on top \\ of a blue cube''} &
%         \scriptsize \makecell{``a corgi wearing a \\ red santa hat''} \\
%         \rule{0pt}{0.2pt}
%     \end{tabular}

%     \caption{Example failures of our pure text-conditional 40M parameter point cloud diffusion model.}
%     \label{fig:textcondfailures}
%     \vskip -0.1in
% \end{figure}

\end{document}